\documentclass{article}

\usepackage{arxiv}                      
\usepackage[numbers,compress]{natbib}   

\usepackage[utf8]{inputenc}
\usepackage[T1]{fontenc}
\usepackage{url}
\usepackage{booktabs}
\usepackage{amsfonts}
\usepackage{amsmath}
\usepackage{amssymb}
\usepackage{amsthm}
\theoremstyle{plain}
\newtheorem{proposition}{Proposition}
\theoremstyle{definition}

\newtheorem{observation}{Observation}
\usepackage{nicefrac}
\usepackage{microtype}
\usepackage{xcolor}
\usepackage{colortbl}
\usepackage{graphicx}
\usepackage{multirow}
\usepackage{tikz}
\usepackage{subcaption}
\usepackage{enumitem}
\usepackage{wrapfig}
\usepackage{placeins} 
\usepackage{tabularx} 
\usepackage{seqsplit} 
\usepackage{hyperref} 

\usetikzlibrary{arrows.meta,positioning,shapes.geometric,fit,backgrounds,calc}

\definecolor{attackred}{HTML}{D32F2F}
\definecolor{defensegreen}{HTML}{388E3C}
\definecolor{stageblue}{HTML}{1565C0}
\definecolor{warnyellow}{HTML}{F57F17}
\definecolor{mydarkblue}{rgb}{0,0.08,0.45}
\hypersetup{
  colorlinks=true,
  linkcolor=mydarkblue,
  citecolor=mydarkblue,
  filecolor=mydarkblue,
  urlcolor=mydarkblue,
  pdfborder={0 0 0}
}

\newcommand{\DTPGD}{49.8}
\newcommand{\RDVLAPGD}{0.0}
\newcommand{\OFTPGD}{18.2}
\newcommand{\DTGauss}{92.7}
\newcommand{\OFTGauss}{89.0}
\newcommand{\RDGauss}{14.8}

\newcommand{\DTOFTp}{0.30}
\newcommand{\KsweepRatio}{1.005}
\newcommand{\KsweepPredicted}{2.0}

\newcommand{\rhoKeight}{48.70}
\newcommand{\rhoKtwelve}{48.96}
\newcommand{\AUCnaive}{0.996}
\newcommand{\AUCadaptive}{0.493}

\newcommand{\amplRatio}{8.22}

\title{Reasoning as a Double-Edged Sword: Architecture and Cross-Stage Robustness in Vision-Language-Action Models}

\renewcommand{\shorttitle}{Reasoning as a Double-Edged Sword in VLAs}

\newcommand{\authorblockN}[1]{#1\\[3pt]}
\newcommand{\authorblockA}[1]{#1}
\author{
\authorblockN{Tuan Duong Trinh\thanks{Corresponding author.},
Naveed Akhtar,
Basim Azam}
\authorblockA{University of Melbourne\\
Parkville, VIC, Australia\\
\{tuanduong.trinh.1, naveed.akhtar1, basim.azam\}@unimelb.edu.au}
}

\begin{document}

\maketitle

\begin{abstract}
Does adding a reasoning step make a Vision-Language-Action (VLA) model more robust to perturbation? Intuitively, a policy that reasons before acting should absorb a perturbed input better than one that maps observations directly to actions. We test this premise head-on across three models that span the reasoning spectrum (no reasoning, a text chain-of-thought, and a latent iterative loop), perturbing each at the vision, reasoning, and action stages on LIBERO and SimplerEnv. Two questions organize the study: does the reasoning design shift robustness, and can the reasoning be read back at runtime as a safety signal? We find that the latent-iterative model is by far the least robust: under both stochastic noise and white-box perturbation its task success collapses, while the other two hold. This fragility is structural rather than cumulative: varying the reasoning depth at inference barely moves it. Reasoning outputs can in principle be monitored, but the monitors fail under fair tests. A plan--action consistency probe that looks near-perfect under naive evaluation falls to chance under adaptive attack. Under matched-FPR calibration, fusing it with an action-anomaly probe never lifts defended success above undefended. Scoped to these output-level behavioral probes under white-box vision-stage attack, this ceiling is a precondition that any viable defense must first satisfy.
   
\end{abstract}

\section{Introduction}
\label{sec:intro}

Vision-Language-Action (VLA) models pair a vision-language backbone with an action head~\citep{brohan2023rt2,kim2024openvla,black2024pi0}, and a growing number add an explicit reasoning step before acting: a chain-of-thought (CoT) plan in text~\citep{zawalski2024ecot,yin2025deepthinkvla}, or iterative refinement in a latent space~\citep{tur2026rdvla}. The safety intuition behind it, that reasoning before acting should absorb a perturbed input better than acting on it directly, to our knowledge, remains largely untested. Existing robustness benchmarks~\citep{fei2025liberoplus} treat the model as a single black box and never isolate the reasoning step, which raises two questions. When a perturbation enters the pipeline, does the reasoning attenuate it, pass it through, or amplify it? And when the reasoning is exposed as an explicit plan, can that plan be read back as a runtime check on the action it drives?

To address these questions, we put the intuition to a direct test. Three VLA models span the reasoning spectrum (Figure~\ref{fig:architecture}): OpenVLA-OFT~\citep{kim2024openvla,kim2025openvlaoft} (no reasoning), DeepThinkVLA~\citep{yin2025deepthinkvla} (text CoT), and RD-VLA~\citep{tur2026rdvla} (latent iterative). We perturb them with a cross-stage attack matrix, stochastic noise and white-box PGD-10~\citep{madry2018pgd} injected at the vision, reasoning, and action stages, across the LIBERO~\citep{liu2023libero} and SimplerEnv~\citep{li2024simpler} manipulation benchmarks.

The latent-reasoning model is the least robust: task success rate (SR) collapses to $\RDGauss\%$ under Gaussian sensor noise and $\RDVLAPGD\%$ under PGD-10, while the other two remain far more robust. Why would a latent reasoning loop amplify a perturbation rather than absorb it? The natural hypothesis is multiplicative compounding with depth, but an inference-time sweep of the recurrence depth $K$ rules it out: extending the recurrence from $K{=}8$ to $K{=}12$ leaves amplification essentially unchanged ($\rho(12)/\rho(8){=}\KsweepRatio$ against a multiplicative prediction of $\KsweepPredicted$). The amplification is structural, not cumulative, a fixed property of the architecture rather than something that grows with depth.

Can the exposed reasoning, then, serve as a defense? A text-reasoning model exposes a plan that a consistency check can read. We evaluate such a check under adaptive attack (standard practice~\citep{carlini2019evaluating}): naive corruption is detected almost perfectly, but adaptive attack reduces it to chance. Fusing this probe with an action-anomaly probe does no better: under matched-FPR calibration, no convex combination of the two raises defended task success above undefended on any PGD-10 cell (\S\ref{sec:intervention}). 

\paragraph{Contributions.}
We make two contributions, one for each question:
\begin{enumerate}[leftmargin=1.4em,itemsep=2pt,topsep=2pt]
\item The fragility we identify is structural, not cumulative. It appears in RD-VLA, our latent-iterative model, whose task success collapses under both Gaussian sensor noise and white-box PGD-10~\citep{madry2018pgd}, while DeepThinkVLA and OpenVLA-OFT remain far more robust. An inference-time $K$-sweep shows depth is not the driver ($\rho(12)/\rho(8){=}\KsweepRatio$, against $\KsweepPredicted$ if it compounded per step), placing the amplifier in the encoder and the recurrence's fixed point.
\item Two behavioral monitors that look promising under naive evaluation fail under fair tests. The plan--action consistency probe collapses to chance under adaptive attack~\citep{carlini2019evaluating} (detection AUC, area under the ROC curve: $\AUCnaive\to\AUCadaptive$), and under matched-FPR calibration no linear fusion of the text-consistency and action-anomaly probes raises defended SR above undefended SR on any PGD-10 cell. We characterize this as a precondition for defense work, not a no-go theorem.
\end{enumerate}

{\setlength{\textfloatsep}{6pt plus 2pt minus 2pt}%
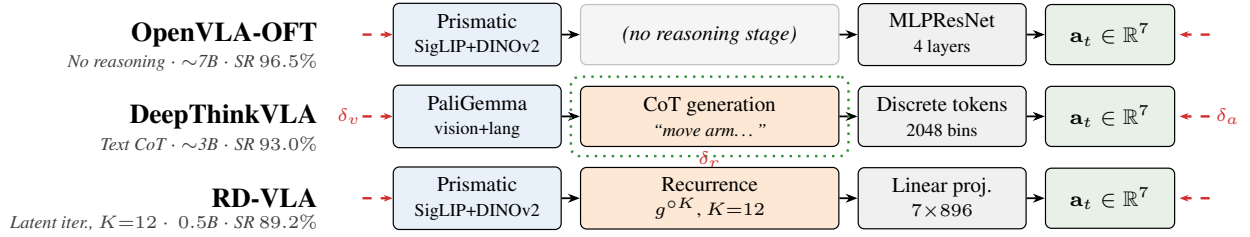
\begin{figure}[t]
\centering
\resizebox{\columnwidth}{!}{%
\begin{tikzpicture}[
  font=\small,
  node distance=6pt and 6pt,
  block/.style={draw, rounded corners=2pt, minimum width=1.95cm, minimum height=0.7cm, align=center, font=\scriptsize},
  vbox/.style={block, fill=stageblue!10},
  rbox/.style={block, fill=warnyellow!18, minimum width=3.0cm}, 
  abox/.style={block, fill=gray!12},
  obox/.style={block, fill=defensegreen!12, minimum width=1.5cm},
  rowlabel/.style={font=\bfseries\small, anchor=east, align=right},
  spec/.style={font=\tiny\itshape, gray!50!black, align=right, anchor=east},
  flow/.style={-{Stealth[length=4pt]}, semithick},
  atk/.style={-{Stealth[length=3pt]}, draw=attackred, dashed, thick},
  atklab/.style={font=\tiny\sffamily, attackred},
  inspect/.style={draw=defensegreen, dotted, thick, rounded corners=3pt}
]

\node[vbox] (oft_v) {Prismatic\\\tiny SigLIP+DINOv2};
\node[rbox, right=of oft_v, fill=gray!8, draw=gray!50] (oft_r) {\scriptsize\itshape (no reasoning stage)};
\node[abox, right=of oft_r] (oft_a) {MLPResNet\\\tiny 4 layers};
\node[obox, right=of oft_a] (oft_o) {$\mathbf{a}_t \in \mathbb{R}^7$};
\draw[flow] (oft_v) -- (oft_r);
\draw[flow] (oft_r) -- (oft_a);
\draw[flow] (oft_a) -- (oft_o);
\node[rowlabel] at ($(oft_v.west)+(-0.75cm,0)$) (oft_t) {OpenVLA-OFT};
\node[spec] at ($(oft_t.south east)+(0,-0.05cm)$) {No reasoning $\cdot\,{\sim}$7B $\cdot$ SR $96.5\%$};

\node[vbox, below=of oft_v] (dt_v) {PaliGemma\\\tiny vision+lang};
\node[rbox, right=of dt_v] (dt_r) {CoT generation\\\tiny\itshape ``move arm\dots''};
\node[abox, right=of dt_r] (dt_a) {Discrete tokens\\\tiny 2048 bins};
\node[obox, right=of dt_a] (dt_o) {$\mathbf{a}_t \in \mathbb{R}^7$};
\draw[flow] (dt_v) -- (dt_r);
\draw[flow] (dt_r) -- (dt_a);
\draw[flow] (dt_a) -- (dt_o);
\node[rowlabel] at ($(dt_v.west)+(-0.75cm,0)$) (dt_t) {DeepThinkVLA};
\node[spec] at ($(dt_t.south east)+(0,-0.05cm)$) {Text CoT $\cdot\,{\sim}$3B $\cdot$ SR $93.0\%$};
\node[inspect, fit=(dt_r), inner sep=3pt] (dt_in) {};

\node[vbox, below=of dt_v] (rd_v) {Prismatic\\\tiny SigLIP+DINOv2};
\node[rbox, right=of rd_v] (rd_r) {Recurrence\\\tiny $g^{\circ K}$, $K{=}12$};
\node[abox, right=of rd_r] (rd_a) {Linear proj.\\\tiny $7{\times}896$};
\node[obox, right=of rd_a] (rd_o) {$\mathbf{a}_t \in \mathbb{R}^7$};
\draw[flow] (rd_v) -- (rd_r);
\draw[flow] (rd_r) -- (rd_a);
\draw[flow] (rd_a) -- (rd_o);
\node[rowlabel] at ($(rd_v.west)+(-0.75cm,0)$) (rd_t) {RD-VLA};
\node[spec] at ($(rd_t.south east)+(0,-0.05cm)$) {Latent iter., $K{=}12 \cdot\,0.5$B $\cdot$ SR $89.2\%$};

\node[atklab, left=7pt of dt_v.west, anchor=east] (vlbl) {$\delta_v$};
\draw[atk] ($(dt_v.west)+(-0.35cm,0cm)$) -- ($(dt_v.west)+(0,0cm)$);
\draw[atk] ($(rd_v.west)+(-0.35cm,0cm)$) -- ($(rd_v.west)+(0,0)$);
\draw[atk] ($(oft_v.west)+(-0.35cm,0cm)$) -- ($(oft_v.west)+(0,0)$);

\node[atklab, below=-2pt of dt_r.south] (rlbl) {$\delta_r$};

\node[atklab, right=10pt of dt_o.east] {$\delta_a$};
\draw[atk] ($(dt_o.east)+(0.4cm,0)$) -- ($(dt_o.east)+(0.05cm,0)$);
\draw[atk] ($(rd_o.east)+(0.4cm,0)$) -- ($(rd_o.east)+(0.05cm,0)$);
\draw[atk] ($(oft_o.east)+(0.4cm,0)$) -- ($(oft_o.east)+(0.05cm,0)$);

\end{tikzpicture}%
}%
\caption{The three VLAs span the reasoning spectrum (rows). \textcolor{red!70!black}{Red dashed arrows} mark stage-wise attack injection points: vision $\delta_v$ (left), reasoning $\delta_r$ (DT only; the only model with a text interface to corrupt), action $\delta_a$ (right). The \textcolor{green!50!black}{green dotted box} highlights DT's text CoT as the only \emph{inspectable} interface available for plan-action consistency probing (\S\ref{sec:adaptive_result}).}
\label{fig:architecture}
\end{figure}}%

\section{Related work}
\label{sec:related}

VLA policies span discrete-token, continuous, and diffusion action heads~\citep{brohan2022rt1,brohan2023rt2,kim2024openvla,kim2025openvlaoft,black2024pi0,black2025pi05,pertsch2025pi0fast,karamcheti2024prismatic,liu2024rdt,octo2024,zhao2023aloha,chi2024diffusion}, trained on large demonstration corpora~\citep{walke2023bridgedata,padalkar2024openx,khazatsky2024droid} and evaluated on simulated manipulation benchmarks~\citep{liu2023libero,mees2022calvin,james2020rlbench,nasiriany2024robocasa,li2024simpler}. A recent line adds an explicit \emph{reasoning} module in one of three modalities: \textbf{text CoT} (Embodied-CoT~\citep{zawalski2024ecot}, ECoT-Lite~\citep{chen2025ecotlite}, DeepThinkVLA~\citep{yin2025deepthinkvla}); \textbf{latent iterative} (RD-VLA~\citep{tur2026rdvla}); \textbf{visual} (CoT-VLA~\citep{zhao2025cotvla}). Our lineup covers the first two plus a single-pass baseline; visual reasoning is out of scope. Reasoning enlarges the attack surface: CoT prompting~\citep{wei2022cot} is exposed to prompt injection~\citep{schulhoff2024hackaprompt}, jailbreaking~\citep{wei2023jailbroken}, GCG suffixes~\citep{zou2023gcg}, and CoT poisoning~\citep{xiang2024badchain,jiang2025safechain}, and CoT (un)faithfulness~\citep{turpin2023unfaithful,lyu2023faithful} undercuts consistency-based defenses~\citep{wang2023selfconsistency,gu2025safe,wu2025steering,li2024cos}. None of this work measures stage-wise amplification of an adversarial perturbation through the reasoning module.

\begin{table}[t]
\centering
\small
\setlength{\tabcolsep}{6pt}
\caption{VLA-safety positioning. V/R/A: vision/reasoning/action; TC/LI/SP: text CoT, latent iterative, single-pass; Adapt.: Carlini-standard eval; Int.: runtime intervention; Arch.: differentiates $\geq 2$ paradigms (\S\ref{sec:limitations}).}
\label{tab:related_work}
\begin{tabular}{@{}lccccc@{}}
\toprule
Work & Stages & Para. & Adapt. & Int. & Arch.\\
\midrule
LIBERO-Plus~\citep{fei2025liberoplus}      & V         & --      & --        & --     & --       \\
TRAP~\citep{huang2026trap}                 & R         & TC      & --        & --     & --       \\
\citet{wang2024vlavulnerabilities}         & V         & SP      & --        & --     & --       \\
\citet{guo2025multimodal}                  & V, R      & TC      & --        & --     & --       \\
\citet{trinh2026altered}                   & R         & TC      & --        & --     & --       \\
\midrule
\textbf{Our work}              & \textbf{V,R,A} & \textbf{TC,LI,SP} & \textbf{$\checkmark$} & \textbf{$\checkmark$} & \textbf{$\checkmark$} \\
\bottomrule
\end{tabular}
\end{table}

Classical adversarial-ML grounds our methodology: gradient-based attacks~\citep{szegedy2014intriguing,goodfellow2015fgsm,kurakin2017bim,carlini2017wagner,madry2018pgd,brown2017advpatch}, adaptive evaluation~\citep{athalye2018obfuscated,carlini2019evaluating,tramer2020adaptive}, gradient-free / transfer / ensemble attacks~\citep{andriushchenko2020square,papernot2016transferability,tramer2017space,croce2020autoattack}, and physical-world threats~\citep{eykholt2018physical}; vision-language models inherit pixel-level vulnerabilities~\citep{zhao2023advvlm,schlarmann2023multimodal}. Certified defenses~\citep{fazlyab2019lipschitz,weng2018lipschitz,cohen2019randomized,wong2018kolter} are typically loose for the attention-heavy backbones in modern VLAs, so we report empirically calibrated per-stage amplification, apply the Carlini adaptive standard, and use Holm--Bonferroni correction with bootstrap CIs for seed variance~\citep{henderson2018reproducibility,agarwal2021precipice}.

\citet{trinh2026altered} (text-CoT entity-swap on DT) is our closest reasoning-stage prior. We extend to all three pipeline stages, three reasoning paradigms, and adaptive evaluation (Table~\ref{tab:related_work}). Narrower-slice priors include TRAP~\citep{huang2026trap} (text-CoT corruption), \citet{wang2024vlavulnerabilities} (single-pass vision attacks), and \citet{guo2025multimodal} (vision + reasoning on a single text-CoT model). LIBERO-Plus~\citep{fei2025liberoplus} is a complementary non-adversarial benchmark sweeping seven robustness axes, most entering at the vision stage.

\section{Setup}
\label{sec:methods}

\subsection{VLA pipeline and threat model}
\label{sec:threat_model}

A VLA model maps visual observations $\mathbf{o}_t \in \mathbb{R}^{H \times W \times 3}$ and a task instruction $\ell$ to a 7-DOF action $\mathbf{a}_t \in \mathbb{R}^7$ through three stages (vision encoder $f_v$, optional reasoning module $f_r$, and action head $f_a$):
\begin{equation}
  \mathbf{a}_t = f_a\bigl(f_r\bigl(f_v(\mathbf{o}_t), \ell\bigr)\bigr).
  \label{eq:pipeline}
\end{equation}
The three models we study (Figure~\ref{fig:architecture}) span the reasoning spectrum: OpenVLA-OFT~\citep{kim2024openvla,kim2025openvlaoft} (single-pass, ${\sim}7$B, Prismatic backbone with MLPResNet action head), DeepThinkVLA (DT; text CoT, ${\sim}3$B, PaliGemma backbone, discrete action tokens), and RD-VLA (latent iterative, $K{=}12$ weight-tied recurrence, $0.5$B, Prismatic backbone, linear action projection). All three are LIBERO-fine-tuned variants for the main-paper experiments (clean SR $96.5\%$ for OFT, $93.0\%$ for DT, $89.2\%$ for RD-VLA) and evaluated on LIBERO~\citep{liu2023libero}'s 4 suites (object, spatial, goal, long-horizon) $\times$ 10 tasks $\times$ 3 seeds; the unit of analysis is the matched (suite, seed) cell, each a mean SR over 50 episodes (10 tasks $\times$ 5 trials), giving $N{=}12$ cells per condition (mean~$\pm$~SE reported throughout). The SimplerEnv cross-benchmark in \S\ref{sec:simpler_env} uses public OpenVLA-7B without LIBERO fine-tuning (clean SR $55.0\%$ on Google-Robot non-floor tasks). The two OpenVLA-family members are different checkpoints with non-comparable absolute SR, so cross-benchmark verdicts use within-row $\Delta$SR (Tab.~\ref{tab:oft_variants}, App.~\ref{app:details}).

To isolate where reasoning architecture absorbs or amplifies a perturbation, we inject one perturbation at a single pipeline stage at a time (Figure~\ref{fig:architecture} marks the three injection points $\delta_v$, $\delta_r$, $\delta_a$). A stage-$s$ adversary with budget $\varepsilon$ perturbs $\tilde{x}_s = x_s + \delta_s$, either deterministically ($\|\delta_s\|_p \leq \varepsilon$, FGSM/PGD-class) or stochastically ($\delta_s \sim \mathcal{N}(0, \sigma^2 I)$), for $s \in \{v, r, a\}$. We perturb the vision (pixel-space $L_\infty$ or Gaussian), reasoning (entity-swap CoT, DT only), and action (Gaussian on the 7-DOF output, gripper excluded) stages (details in \S\ref{sec:attacks}). We do not perturb the natural-language instruction $\ell$ itself: instruction-level attacks (paraphrase, prompt injection) form a separate threat class that does not isolate reasoning-architecture-mediated effects, and our reasoning-stage entity swap probes a related semantic-content corruption at the model's \emph{internally generated} plan rather than the input. All quantities downstream of the vision stage flow through the same perturbed latent $\tilde{\mathbf{z}}_t = f_v(\tilde{\mathbf{o}}_t)$. This shared dependency underpins Observation~\ref{obs:coherent} (App.~\ref{app:coherent_bound}).

\subsection{Cross-stage attack matrix and physical analogs}
\label{sec:attacks}
\label{sec:threat_mapping}

\textbf{Vision stage.}
All three models share an identical FGSM~\citep{goodfellow2015fgsm} sweep at $\varepsilon\in\{2,4,8,16,32\}/255$ and Gaussian sweep at $\sigma\in\{0.01,0.02,0.05,0.1,0.2\}$. PGD-10~\citep{madry2018pgd} ($L_\infty$, step $\varepsilon/4$; $\varepsilon{=}8/255$ is the canonical $L_\infty$ budget in the adversarial-ML literature, and our headline cross-model tables report PGD-10 at this operating point) is run at $\varepsilon\in\{4,8,16\}/255$ for DT and at $\varepsilon{=}8/255$ for OFT only, restricted because OFT PGD-10 costs ${\sim}10\times$ FGSM compute. White-box PGD through RD-VLA's $K{=}12$ recurrence is enabled by activation checkpointing; RD-VLA FGSM cells are reported as DT$\to$RD-VLA transfer attacks~\citep{papernot2016transferability,tramer2017space} (a documented white-box lower bound) since PGD-10 strictly dominates FGSM on the same model. We complement with the gradient-free Square Attack~\citep{andriushchenko2020square} at 100 queries. Per-architecture loss functions, gradient details, and the activation-checkpointing implementation: App.~\ref{app:details}. \textbf{Reasoning stage (DT only).}
Entity-swap CoT corruption~\citep{trinh2026altered}: the task-relevant object in the model's generated CoT is replaced with a random alternative from LIBERO's 29-object vocabulary. We use this attack because it is the most semantically minimal corruption that inverts the action target while preserving syntactic well-formedness, directly testing whether the action head is causally coupled to the CoT's entity field. Direction- and gripper-state swaps probe orthogonal dimensions of CoT-action coupling and are deferred to future work. Inapplicable to RD-VLA (no text) and OFT (no reasoning). \textbf{Action stage.}
Additive Gaussian noise on the 7-DOF output (gripper excluded), $\sigma\in\{0.01,0.05,0.1,0.5,1.0\}$, identical across all three models.

\paragraph{Physical analogs.}
These analogs are class-level rather than calibrated. The cited work motivates that the corresponding noise classes arise in real-robot deployment, while our $\sigma$ and $\varepsilon$ sweeps isolate stage-local sensitivity rather than reproducing any specific camera, actuator, or text-channel process. \textbf{Vision-stage analogs.} Additive Gaussian RGB noise is a stage-local proxy for stochastic imaging degradation such as sensor read noise, ISO grain, and lighting variation. Related common corruptions are catalogued in ImageNet-C~\citep{hendrycks2019corruptions}, and sim-to-real work randomizes appearance and illumination factors because such variation affects real transfer~\citep{tobin2017domainrand}. LIBERO-Plus~\citep{fei2025liberoplus} documents modern VLAs collapsing under this noise class. Norm-bounded RGB attacks (FGSM, PGD-10) are not calibrated models of physical adversarial patches or printed perturbations~\citep{brown2017advpatch,eykholt2018physical}. We include them as worst-case white-box diagnostics of visual-pathway sensitivity, complementary to the Gaussian condition. Failure under PGD should be read as evidence of small-margin visual dependence, not as evidence that the same perturbation physically transfers. \textbf{Action-stage analogs.} Additive Gaussian noise on the 7-DOF output is a stage-local proxy for command-to-motion mismatch arising from actuator slip, encoder drift, and end-effector miscalibration. Dynamics-randomization work motivates this noise class as a deployment concern~\citep{peng2018dynamics}. We do not claim our output-space Gaussian reproduces the randomization protocol studied there. \textbf{Reasoning-stage analogs.} Entity swaps in the textual CoT are a controlled semantic corruption of the reasoning channel, analogous to upstream language-channel or grounding errors such as OCR/ASR substitutions or incorrect object-name binding. Prior work shows that neural sequence models can be brittle to text-channel noise~\citep{belinkov2018noise}. This analog applies only to the text-CoT model (DT). Physical-world transfer itself remains future work (\S\ref{sec:limitations}).

\section{Cross-stage robustness across reasoning architectures}
\label{sec:crossstage}

Does the reasoning stage soften a perturbation, relay it, or magnify it? We locate the answer across the three models (\S\ref{sec:vulnerability}), then trace the fragility to its architectural source (\S\ref{sec:ksweep_result}).

\subsection{Cross-stage robustness ordering}
\label{sec:vulnerability}

\begin{figure}[t]
  \centering
  \includegraphics[width=\textwidth]{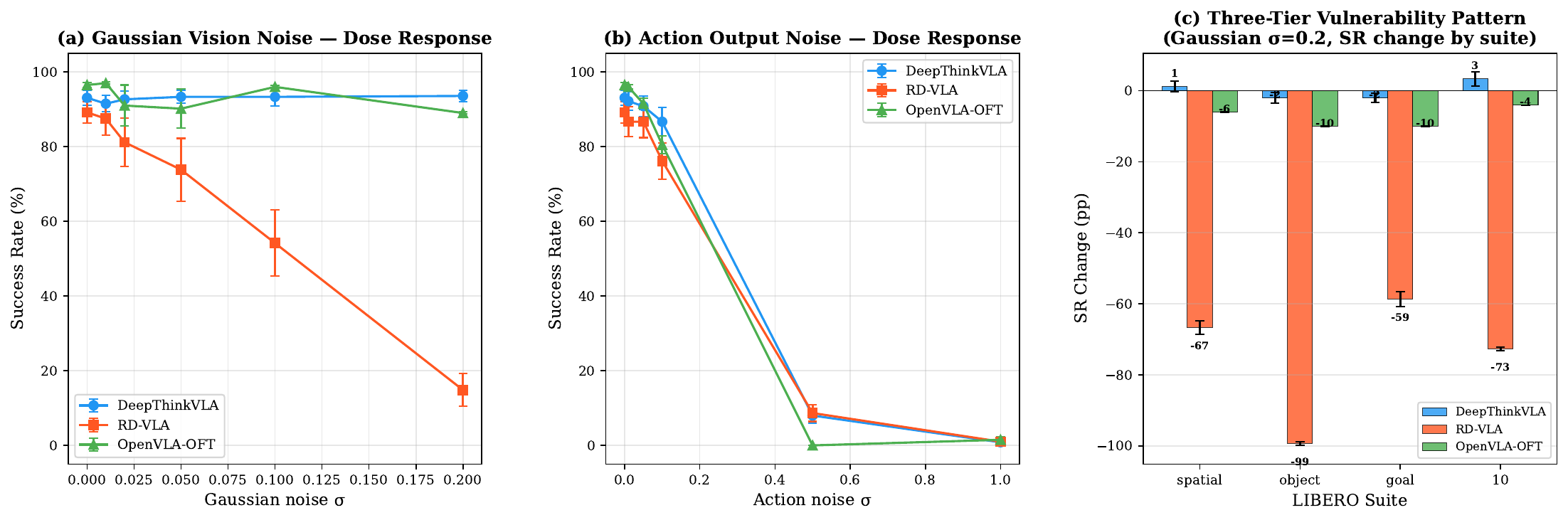}
  \caption{Cross-stage vulnerability profile. \textbf{(a)} Gaussian vision noise dose-response: DT and OFT withstand, RD-VLA collapses. \textbf{(b)} Action-stage Gaussian noise produces a sharp cliff between $\sigma{=}0.1$ and $\sigma{=}0.5$ on all three models. \textbf{(c)} Two-tier Gaussian vulnerability at $\sigma{=}0.2$ ($\Delta$SR by suite): RD-VLA collapses ($-74.3$pp) while DT ($-0.3$pp) and OFT ($-7.5$pp) are inconclusive at $N{=}12$ ($p{=}\DTOFTp$). Error bars $\pm 1$ SE. The FGSM dose-response is in App.~\ref{app:eps_std}; the PGD-10 row, which separates DT from OFT, is in Table~\ref{tab:headline_3x3}.}
  \label{fig:cross_stage}
\end{figure}

We evaluate vulnerability across 18 stage-wise attack conditions on 4 LIBERO suites $\times$ 3 seeds per (model, condition) cell ($N{=}12$ per cell). Figure~\ref{fig:cross_stage} presents the cross-stage profile, Table~\ref{tab:headline_3x3} the canonical-operating-point comparison; full per-condition statistics in App.~\ref{app:per_suite}.

\begin{table}[t]
\centering
\small
\setlength{\tabcolsep}{6pt}
\caption{Cross-stage robustness summary. DT and OFT withstand Gaussian noise where RD-VLA collapses. PGD-10 separates DT from OFT and establishes DT~$>$~OFT~$\gg$~RD-VLA at $\varepsilon{=}8/255$ (DT~$\DTPGD\%$ vs.\ OFT~$\OFTPGD\%$ SR; Gaussian DT-vs-OFT inconclusive). Cells: $^{w}$ white-box; $^{t}$ DT$\to$RD-VLA transfer (FGSM only); $^{s}$ stochastic. Cell shading encodes SR: \textcolor{green!50!black}{green} $\geq 80\%$, \textcolor{yellow!60!black}{yellow} $50$--$80\%$, \textcolor{orange!80!black}{orange} $20$--$50\%$, \textcolor{red!70!black}{red} $<20\%$. App.~\ref{app:eps_std} extends to $\varepsilon{\in}\{4,8\}/255$.}
\label{tab:headline_3x3}
\begin{tabular}{@{}lccc}
\toprule
Attack & OFT & DT & RD-VLA \\
\midrule
Gaussian $\sigma{=}0.2$ & \cellcolor{green!25}$89.0^{s}$ & \cellcolor{green!25}$92.7^{s}$ & \cellcolor{red!30}$14.8^{s}$ \\
FGSM $\varepsilon{=}8/255$ & \cellcolor{green!25}$83.5^{w}$ & \cellcolor{yellow!25}$55.8^{w}$ & \cellcolor{yellow!25}$65.2^{t}$ \\
PGD-10 $\varepsilon{=}8/255$ & \cellcolor{red!30}$\OFTPGD^{w}$ & \cellcolor{orange!25}$\DTPGD^{w}$ & \cellcolor{red!50}$\RDVLAPGD^{w}$ \\
\bottomrule
\end{tabular}
\end{table}

\begin{figure}[t]
\centering
\includegraphics[width=\textwidth]{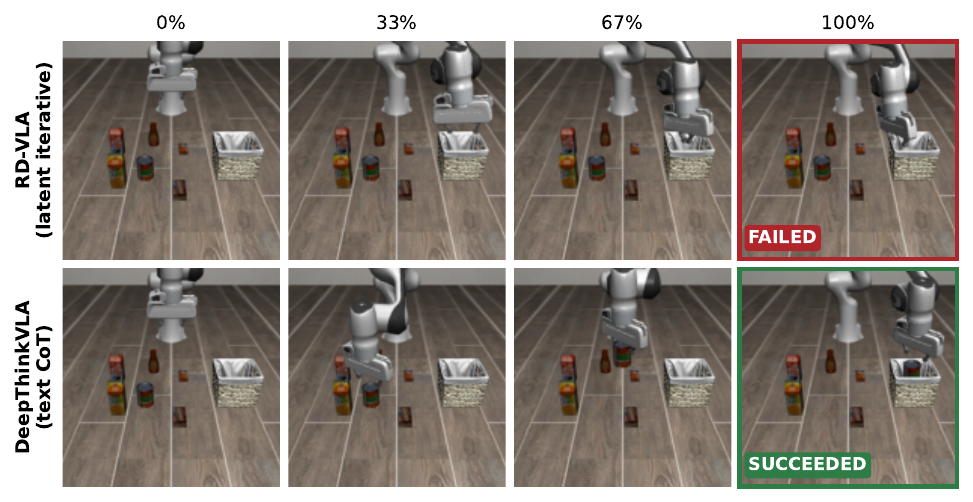}
\caption{\textbf{The $\sigma{=}0.2$ collapse, in rollouts.} Representative episodes under Gaussian sensor noise ($\sigma{=}0.2$) on a \texttt{libero\_object} task (``pick up the tomato sauce and place it in the basket''), sampled at even fractions ($0\%$--$100\%$) of each rollout from an identical start state. The latent-iterative RD-VLA (top) leaves the object untouched and \textbf{fails}, while the text-CoT DeepThinkVLA (bottom) completes the pick-and-place and \textbf{succeeds}, the per-episode face of the aggregate gap ($\RDGauss\%$ vs.\ $\DTGauss\%$ SR, $N{=}12$; Table~\ref{tab:headline_3x3}). These are single illustrative episodes; the quantitative claims rest on the $N{=}12$ aggregates. Full rollouts are in the supplementary video.}
\label{fig:qualitative_collapse}
\end{figure}

Under Gaussian noise at $\sigma{=}0.2$, RD-VLA collapses ($\RDGauss\%$, $\Delta{=}{-}74.3$pp) while DT~($\DTGauss\%$) and OFT~($\OFTGauss\%$) withstand corruption (Fig.~\ref{fig:qualitative_collapse}). The Gaussian ordering is two-tier: the DT-vs-OFT gap is inconclusive at $N{=}12$ ($\Delta_{\text{DT-OFT}}{=}{+}3.7$pp, $d{=}0.87$, paired Wilcoxon signed-rank test, the non-parametric paired test for matched (suite, seed) cells, with Holm-Bonferroni correction over the 13-test cross-model family, $p{=}\DTOFTp$), and both DT and OFT differ from RD-VLA significantly ($p{<}0.01$, $d{=}6.5$). White-box PGD-10 at $\varepsilon{=}8/255$ separates DT from OFT and establishes a strict three-tier ordering: DT~$\DTPGD\%$, OFT~$\OFTPGD\%$, RD-VLA~$\RDVLAPGD\%$ (white-box on all three, RD-VLA via gradient checkpointing through the $K{=}12$ recurrence). To rule out the action stage as the driver, we also perturb there directly: Gaussian noise at the action stage produces a sharp cliff between $\sigma{=}0.1$ ($\geq 76\%$ SR) and $\sigma{=}0.5$ ($\leq 9\%$) on all three models, consistent with action attacks bypassing the reasoning module entirely~\citep{guo2025multimodal}. Only under FGSM does OFT reorder above DT ($83.5\%$ vs $55.8\%$ at $\varepsilon{=}8/255$, a $27.7$pp gap), which we analyze below.

\paragraph{Mechanism: gradient masking refuted.}
\label{sec:eot_mechanism}
Our EOT~\citep{athalye2018obfuscated} test at $T{=}2$ narrows the OFT FGSM-vs-PGD-10 gap to ${\leq}4$pp on 3/4 suites and strengthens PGD-10 slightly on average ($\Delta_\text{PGD}{=}{-}1.7$pp), and gradient-free Square Attack at 100 queries leaves OFT at $95.5\%$ SR. The residual asymmetry is consistent with loss-surface curvature on OFT's MLPResNet head, not gradient masking (App.~\ref{app:eot_mechanism}; Hessian-trace probe and AutoAttack~\citep{croce2020autoattack} are future work).

\subsection{\texorpdfstring{$K$}{K}-sweep falsification: amplification is \texorpdfstring{$K$}{K}-invariant, not multiplicative}
\label{sec:ksweep_result}

To explain RD-VLA's vulnerability, we test the standard mechanistic intuition for weight-tied recurrent reasoning. RD-VLA's reasoning module is a recurrence $f_r = g^{\circ K}$: the same function $g$ applied $K{=}12$ times at inference. The end-to-end amplification ratio $\rho(K)$ measures how much an input perturbation is magnified at the output, per unit input. The standard Lipschitz intuition (Lipschitz: an upper bound on per-layer sensitivity) predicts multiplicative growth, $\rho(K) \approx L_\text{iter}^K$, so a deeper loop should be the more fragile one (the formal three-stage Lipschitz bound is in App.~\ref{app:propagation_bounds}). We test directly by varying $K$ at inference. Running $K \in \{4, 8, 12\}$ on the same evaluation episodes yields $\rho(12)/\rho(8){=}\KsweepRatio$ (Table~\ref{tab:e7_ksweep}). The multiplicative prediction at the back-solved $\hat{L}_\text{iter}{=}1.192$ is $\rho(12)/\rho(8){=}\KsweepPredicted$. The observed $\rho(12)$ sits ${\sim}80$ pooled-SE units below this prediction. The multiplicative form is falsified.

\begin{proposition}[Recurrence amplification is structural and in-distribution $K$-invariant]\label{prop:recurrence}
For weight-tied recurrence $f_r = g^{\circ K}$, the observed end-to-end amplification ratio $\rho_\text{RD-VLA}/\rho_\text{OFT}{=}\amplRatio$ at $\sigma{=}0.2$ is consistent with the reasoning stage as the dominant amplifier and is $K$-invariant in the in-distribution range $K{\in}\{8,12\}$ ($\rho(8){=}\rhoKeight$, $\rho(12){=}\rhoKtwelve$, SE${\leq}0.6$, $N{=}12$ per $K$; the OOD $K{=}4$ point yields the same pattern, App.~\ref{app:amplification}). The multiplicative $L_\text{iter}^K$ form is falsified ($\rho(12)/\rho(8){=}\KsweepRatio$ vs predicted $\KsweepPredicted$). A stage-level Lipschitz claim isolating encoder, reasoning, and action-head contributions is future work.
\end{proposition}

\begin{table}[ht]
\centering
\caption{RD-VLA $K$-sweep at $\varepsilon{=}8/255$ ($N{=}12$ per $K$, all 4 LIBERO suites $\times$ 3 seeds). Per-step amplification $\rho(K) = \overline{\|\Delta a\|_2}/\varepsilon$ is statistically flat across $K$, falsifying the $L_\text{iter}^K$ hypothesis (which would predict $1.192^K$, growing $4{\times}$ from $K{=}4$ to $K{=}12$). Implied $\hat L_\text{iter}$ is $\rho(12)/\rho(K))^{1/(12-K)}$; values $\approx 1.0$ indicate no per-iteration amplification. Free-intercept fit: $\rho(K) = 48.507 \cdot 1.0007^K$, MAFE $0.1\%$; paper back-solve baseline was $1.192$. Caveat: $K{=}4$ clean SR is $72.8\%$ vs.\ $89\%$ at $K{=}12$, indicating $K{=}4$ is mildly OOD (model trained at $K{=}12$); the $K{\in}\{8,12\}$ comparison alone (both clean SR ${\geq}88\%$, ratio $\rho(12)/\rho(8){=}1.005$) is sufficient to falsify $L_\text{iter}{=}1.192$.}
\label{tab:e7_ksweep}
\begin{tabular}{rrrrr}
\toprule
$K$ & Clean SR (\%) & Attacked SR (\%) & $\rho(K)$ & $\hat L_\text{iter}$ implied \\
\midrule
4 & 72.8 $\pm$ 6.6 & 0.0 $\pm$ 0.0 & 48.682 $\pm$ 0.619 & 1.0007 \\
8 & 88.0 $\pm$ 4.2 & 0.0 $\pm$ 0.0 & 48.702 $\pm$ 0.409 & 1.0013 \\
12 & 89.2 $\pm$ 3.0 & 0.0 $\pm$ 0.0 & 48.956 $\pm$ 0.297 & --- \\
\bottomrule
\end{tabular}
\end{table}

$K$-invariance suggests the dominant amplifier is the encoder plus the recurrence's fixed-point output rather than the per-iteration accumulation. The $K{=}4$ point is out-of-distribution (OOD; RD-VLA was trained at $K{=}12$). We retain it because the OOD-stability is itself informative: the per-iteration multiplicative form does not materialise even when the recurrence is pushed off its trained operating point. The $K$-invariance claim therefore rests on the in-distribution range $K{\in}\{8,12\}$. In DT, under FGSM, the CoT text diverges from clean (edit distance $0.37$--$0.43$, semantic cosine $0.90$--$0.92$) and the magnitude of that divergence correlates with the resulting action deviation (DT under FGSM: Pearson $r{=}{+}0.44$, $N{=}3{,}000$, $p{<}0.0001$; larger CoT drift goes with larger action drift; full amplification cascade in App.~\ref{app:amplification}). The pattern is consistent with a three-stage cascade $\delta_v \to \delta_\text{CoT} \to \delta_\mathbf{a}$, although the single correlation does not establish the directed-graph structure.

\subsection{Confound analysis and external-validity checks}
\label{sec:confounds}

\textbf{We attribute the cross-architecture pattern to reasoning type with four controls}, ruling out backbone, scale, action head, and training data (full analysis App.~\ref{app:confounds}). (i) RD-VLA and OFT share the Prismatic backbone but differ at the reasoning stage ($K{=}12$ weight-tied recurrence vs.\ single-pass) and differ by $9.9{\times}$ at $\sigma{=}0.2$: backbone alone cannot explain the gap. (ii) The SR-vulnerability ordering (DT~$<$~OFT~$<$~RD-VLA) is non-monotone in parameters and per-step amplification, ruling out scale and downstream sensitivity. (iii) RD-VLA's output projection is a contraction ($\sigma_1{=}0.091$) and DT's discrete tokenization is a quantization barrier at the action head: the amplification originates upstream, in the recurrent reasoning layers (Prop.~\ref{prop:recurrence}). (iv) A CoT-disabled ablation on DT yields no detected CoT-on advantage at $\sigma{=}0.2$ ($\Delta{=}{+}1.5$pp, paired $t$, $N{=}12$, $p{=}0.35$). Equivalence tests place the CoT effect within $\pm 5$pp, with the caveat that the zero-token control is off-manifold relative to an in-distribution skip-CoT comparator (future work). Read together, the four controls localise the within-Prismatic recurrence as the load-bearing architectural attribution behind the $9.9{\times}$ SR-degradation gap (RD-VLA vs.\ OFT at $\sigma{=}0.2$). The small DT-OFT residual is consistent with backbone rather than CoT.

\label{sec:simpler_env}\textbf{SimplerEnv (cross-architecture sanity).} On SimplerEnv~\citep{li2024simpler}, the OFT FGSM/PGD-10 asymmetry replicates on the public OpenVLA-7B checkpoint (PGD-10 exceeds FGSM on every task by mean $17.2$\,pp). Only OpenVLA-7B has a Bridge/Fractal-compatible checkpoint, so we treat this as a single-architecture sanity check rather than cross-model replication (per-task details and viability-exclusion notes in App.~\ref{app:simpler_env}). \label{sec:libero_plus}\textbf{LIBERO-Plus (cross-class consistency).} On LIBERO-Plus~\citep{fei2025liberoplus}'s seven naturalistic-perturbation factors (not a $K$-sweep test of Prop.~\ref{prop:recurrence}), we find the RD-VLA-vs-\{DT, OFT\} paired $\Delta$SR gap exceeds $30$\,pp on six of seven factors. The exception is robot initial state, where the RD-vs-OFT gap collapses to $+9.6$\,pp and all three policies degrade comparably ($\Delta\mathrm{SR}{=}{-}57.5$ vs ${-}67.1$\,pp), so the structural-amplifier reading does not extend to starting-state shifts (per-factor results and selection-bias diagnostic in App.~\ref{app:libero_plus}).

\section{Reasoning-based runtime defenses}
\label{sec:defenses}

The reasoning stage drives how a perturbation propagates (\S\ref{sec:crossstage}). Can it also supply a runtime safety signal? We probe two behavioral monitors under an adaptive attacker: a plan--action consistency probe (\S\ref{sec:adaptive_result}) and a stage-fused monitor (\S\ref{sec:intervention}), calibrated at matched FPR across all fusion weights.

\subsection{Adaptive consistency-probe ceiling}
\label{sec:adaptive_result}

The plan--action consistency probe (text CoT only) scores plan--instruction agreement on the CoT:
\begin{equation}
  s = w_e S_\text{entity} + w_d S_\text{direction} + w_g S_\text{gripper} + w_p S_\text{parse},
  \label{eq:consistency}
\end{equation}
with $S_\text{entity}$ scoring the match between the CoT's object reference and the instruction's, $S_\text{direction}$ the motion direction, $S_\text{gripper}$ the grasp state, and $S_\text{parse}$ syntactic well-formedness. Form and weights $(0.5, 0.2, 0.15, 0.15)$ follow~\citet{trinh2026altered}: weights are not load-bearing because the entity term carries the signal (matching details: App.~\ref{app:checker}; per-component AUCs across the ladder: App.~\ref{app:adaptive}). Following~\citet{carlini2019evaluating}, the probe is evaluated against an attacker who knows it and adapts. The ladder uses the same entity-swap corruption at every rung but varies what the attacker hides from the probe: (i) naive (no evasion); (ii) direction-fix (corrects parse and direction features); (iii) stealth entity re-injection (reinserts the instruction's true entity into the corrupted CoT); (iv) full adaptive (both). All four rungs produce the same task disruption; only AUC shifts. We treat the probe diagnostically, not as a proposed defense.

The probe's AUC against naive entity-swap is $0.996$. It collapses to $0.493$ (chance) under stealth entity re-injection and to $0.259$ under full adaptive (Table~\ref{tab:detection}). In a live end-to-end run, stealth disrupts the task as much as naive entity swap ($-8$ pp SR) but is essentially invisible to the probe: its flag rate falls from $98.9\%$ to $5.5\%$, indistinguishable from the $6.3\%$ clean false-positive rate (FPR). We read this as an empirical ceiling for entity-matching probes, not a structural no-go theorem for consistency-based detection. Vision-stage attacks need no adaptive ladder to defeat the probe: detection AUC is already near chance under static FGSM and Gaussian noise (AUC~$\leq 0.692$ across all FGSM levels; AUC~$=0.463$ for Gaussian $\sigma{=}0.2$; Table~\ref{tab:detection}). The reason is structural: vision-stage perturbations enter both branches through the same perturbed latent, so a consistency probe sees both shift coherently. The clean-vs-attacked mean separation is bounded above by $1-\bar\phi_\mathrm{clean}$ (formal statement: App.~\ref{app:coherent_bound}). Empirically, $\bar\phi_\mathrm{clean}{=}0.867{\pm}0.025$ on $N{=}200$ DT episodes (bound $1-\bar\phi_\mathrm{clean}{=}0.133$), and the clean-vs-attacked shift in mean is $\leq 0.022$ across all FGSM/Gaussian levels, well inside the bound (per-suite bootstrap in App.~\ref{app:phi_clean_bootstrap}).

\begin{table}[t]
\centering
\small
\setlength{\tabcolsep}{3pt}
\renewcommand{\arraystretch}{0.95}
\caption{Consistency probe separation by attack type and stage. AUC-ROC, one-sided (lower $=$ more suspicious). Text attacks: C5-Ph1 CoT corpus ($N{=}300$/cond.); vision attacks: DT episodes. Detected? $=$ AUC$>$0.7 and TPR@10\%$>$50\%.}
\label{tab:detection}
\resizebox{\ifdim\width>\textwidth \textwidth\else\width\fi}{!}{%
\begin{tabular}{@{}llcccc@{}}
\toprule
Attack & Stage & AUC & TPR@5\% & TPR@10\% & Detected? \\
\midrule
  Entity swap & Text & $0.996$ & $100.0\%$ & $100.0\%$ & \checkmark \\
  Adaptive stealth & Text & $0.493$ & $1.7\%$ & $6.7\%$ & $\times$ \\
  Adaptive full & Text & $0.259$ & $1.3\%$ & $1.3\%$ & $\times$ \\
  Garbled 100\% & Text & $0.999$ & $100.0\%$ & $100.0\%$ & \checkmark \\
  Garbled 50\% & Text & $0.788$ & --- & $59.5\%$ & \checkmark \\
  Shuffled & Text & $0.498$ & $10.0\%$ & $10.0\%$ & $\times$ \\
\midrule
  FGSM $\varepsilon{=}2/255$ & Vision & $0.673$ & --- & $28.0\%$ & $\times$ \\
  FGSM $\varepsilon{=}4/255$ & Vision & $0.617$ & --- & $24.0\%$ & $\times$ \\
  FGSM $\varepsilon{=}8/255$ & Vision & $0.551$ & --- & $19.0\%$ & $\times$ \\
  FGSM $\varepsilon{=}16/255$ & Vision & $0.692$ & --- & $36.0\%$ & $\times$ \\
  FGSM $\varepsilon{=}32/255$ & Vision & $0.637$ & --- & $38.0\%$ & $\times$ \\
  Gaussian $\sigma{=}0.05$ & Vision & $0.625$ & --- & $20.0\%$ & $\times$ \\
  Gaussian $\sigma{=}0.2$ & Vision & $0.463$ & --- & $8.5\%$ & $\times$ \\
\bottomrule
\end{tabular}%
}
\end{table}

\subsection{Stage-fused monitor ceiling under matched FPR}
\label{sec:intervention}

The consistency probe alone is bounded by the coherent-shift mechanism (Obs.~\ref{obs:coherent}). Under a vision-stage attack, the consistency signal and the action drift together. We therefore add an action-stage signal. The guard scores each step
\begin{equation}
  s_t = \alpha\,(1-\phi_t) + (1-\alpha)\,\psi_t, \quad \alpha\in[0,1],
  \label{eq:guard}
\end{equation}
where $\phi_t$ is the consistency probe (Eq.~\ref{eq:consistency}) and $\psi_t \in [0,1]$ is the empirical-CDF rank of the Mahalanobis distance $d_k(\mathbf{a}_t)$ among clean rollouts for task $k$. A step is flagged when $s_t > \tau$. The threshold $\tau$ is calibrated leave-one-seed-out to an $8\%$ clean episode-level false-positive rate, identically at every $\alpha$ in the sweep. Evaluation runs post hoc on the existing trajectory corpus under an abstain-on-flag rule, so defended SR is bounded above by undefended SR by construction. Even without an attack the bound is active: at $\alpha{=}0.5$ the guard abstains on enough successful clean rollouts to cost ${\sim}8$\,pp of clean task success on average ($10.7$ and $10.0$\,pp on \texttt{libero\_object} and \texttt{libero\_spatial}; App.~\ref{app:alpha_sweep}).

\begin{table}[t]
\centering
\small
\setlength{\tabcolsep}{5pt}
\caption{Pooled $\alpha$-sweep of the stage-fused monitor on DeepThinkVLA (4 LIBERO suites, $N{=}600$ per cell, PGD-10 $\varepsilon{=}8/255$). Threshold $\tau$ calibrated leave-one-seed-out to a target $8\%$ clean FPR (realized shown). Under abstain-on-flag, Defended SR counts task success with no flag, so Defended SR $\leq$ Raw SR. $\Delta = $ Defended $-$ Raw (pp), $95\%$ paired-bootstrap CI ($n{=}10000$). \textbf{Bold}: best-case setting ($\alpha{=}1$, text-only). \underline{Underline}: the stage-fused monitor ($\alpha{=}0.5$). Per-suite and Gaussian breakdowns in App.~\ref{app:alpha_sweep}.}
\label{tab:alpha_sweep}
\begin{tabular}{cccccc}
\toprule
 & Clean & \multicolumn{4}{c}{PGD-10 ($\varepsilon{=}8/255$)} \\
\cmidrule(lr){2-2}\cmidrule(lr){3-6}
$\alpha$ & FPR (\%) & Raw SR (\%) & Defended SR (\%) & $\Delta$ (pp) & Trigger (\%) \\
\midrule
$0$ & $3.5$ & $49.8$ & $42.2$ & $-7.7$~{\scriptsize$[-9.8, -5.7]$} & $19.8$ \\
$0.25$ & $9.5$ & $49.8$ & $36.7$ & $-13.2$~{\scriptsize$[-16.0, -10.5]$} & $40.8$ \\
$0.5$ & $8.5$ & $49.8$ & $\underline{42.2}$ & $\underline{-7.7}$~{\scriptsize$[-9.8, -5.7]$} & $27.7$ \\
$0.75$ & $9.0$ & $49.8$ & $42.8$ & $-7.0$~{\scriptsize$[-9.2, -5.0]$} & $23.7$ \\
$1$ & $3.7$ & $49.8$ & $\mathbf{48.8}$ & $\mathbf{-1.0}$~{\scriptsize$[-1.8, -0.3]$} & $4.2$ \\
\bottomrule
\end{tabular}
\end{table}

The matched-FPR sweep maps the ceiling along the fusion-weight axis. Defended task success stays below undefended at every $\alpha$ (Table~\ref{tab:alpha_sweep}; per-suite and bootstrap CIs in App.~\ref{app:alpha_sweep}). The calibrated text-only baseline ($\alpha{=}1$) recovers $48.8\%$ mean defended SR. The calibrated stage-fused monitor ($\alpha{=}0.5$) recovers $42.2\%$. Fusion is worse than its own text-only ablation. The action-only component ($\alpha{=}0$) flags every white-box PGD-10 episode on RD-VLA, but RD-VLA's undefended SR is already $0\%$. Detection transfers across architectures, but task utility does not. This is an $\alpha$-axis ceiling: no convex combination of these two probe families raises defended SR above undefended SR under fair-FPR calibration on this threat model.

Without matched-FPR calibration, the same monitor pair gives the opposite result. With $\tau{=}0.7$ on the text-only baseline and no leave-one-seed-out calibration, the fused monitor at $\alpha{=}0.5$ shows a $+24.6$\,pp pooled improvement. The $\tau{=}0.7$ baseline abstains on $40$--$67\%$ of clean episodes, five to eight times the $8\%$ target. The $+24.6$\,pp gap is the baseline's miscalibration, not the fused monitor's contribution. We exclude this comparison from the headline.

\begin{wrapfigure}{r}{0.5\textwidth}
  \centering
  \includegraphics[width=\linewidth]{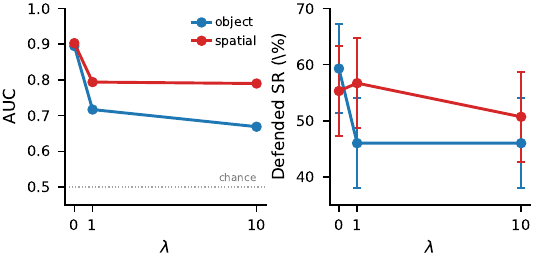}
  \caption{Stage-fused guard ($\alpha{=}0.5$) under checker-aware adaptive PGD-10 ($\varepsilon{=}8/255$). As the action-anomaly penalty weight $\lambda$ increases, both detection AUC (left) and defended SR (right) drop on \texttt{libero\_object} and \texttt{libero\_spatial}. Numerical values: App.~\ref{app:I3_table}.}
  \label{fig:I3_lambda_sweep}
\end{wrapfigure}
An adaptive attacker drives defended SR lower still. We attack the guard with a checker-aware PGD-10 whose loss adds a differentiable penalty $-\lambda\,d_k(\hat{\mathbf{a}}_t)^2$ on a soft-argmax estimate of the predicted action, swept over $\lambda \in \{0,1,10\}$. Both detection AUC and defended SR fall as $\lambda$ increases (Figure~\ref{fig:I3_lambda_sweep}). Worst-case defended SR drops to $46.0\%$ on object and $50.7\%$ on spatial, both well below undefended on the same attack. The attack is white-box on the action-anomaly signal only. Jointly adapting against the text-consistency term would loosen the bound further (\S\ref{sec:limitations}). Transfer to OpenVLA-OFT is future work, as the available rollouts store only aggregate outcomes, not the per-step $\psi_t$ the guard requires. Runtime cost is ${\sim}0.01$\,ms per control step, negligible against per-step VLA inference.

\section{Discussion and conclusion}
\label{sec:discussion}

We do not propose a defense. Our results establish a ceiling for two widely-studied behavioral post-hoc probe families (text-consistency on the reasoning stage, Mahalanobis action-anomaly on the action stage): under Carlini-standard adaptive attack and fair-FPR calibration, no convex fusion raises defended task success above undefended success on any PGD-10 cell (\S\ref{sec:intervention}). We characterize this ceiling along two axes (probe family, fusion weight) and view it as a precondition for defense work --- not a no-go theorem. Both probes read stage outputs. Three directions lie beyond the scope we tested: probes that access intermediate representations rather than their outputs, training-time interventions on the reasoning stage, and threat models that exclude white-box vision-stage attack. Whether any of these directions breaks the documented ceiling is an open question.

Two architecture-specific observations follow. (1) RD-VLA, our latent-iterative model, is the most vulnerable of the three: the largest amplification ($-74$pp under Gaussian) and no inspectable plan a text-based monitor could read. This amplifier reading is specific to perturbation classes that engage the recurrence: on LIBERO-Plus's robot-initial-state factor all three models degrade comparably and the RD-vs-OFT gap collapses to $+9.6$pp (\S\ref{sec:libero_plus}, App.~\ref{app:libero_plus}). Plausible mitigations (convergence monitoring; learned latent-trajectory probes) need within-paradigm replication to move beyond conjecture. (2) Text CoT's measurable contribution to safety is inspectability: the probe in \S\ref{sec:adaptive_result} exists because there is a plan to read. A noise-filtering contribution is below our power threshold ($N{=}12$, minimum detectable effect $d{\approx}0.85$); the natural test is a higher-$N$ replication. We claim inspectability is the measurable contribution, not the primary one. Across our three models, what separates them under cross-stage attack is how the reasoning stage is built, not whether one is present.

\section{Limitations}
\label{sec:limitations}

Our findings come with three limitations that open avenues for future work. Each reasoning paradigm is represented by a single model ($N{=}3$), so backbone, scale, training data, and action head are confounded with reasoning family. Every paradigm-level statement is therefore a model-level observation, and within-paradigm replication is a natural extension (App.~\ref{app:limitations}). Consistent with this, the LIBERO-Plus cross-class consistency check (\S\ref{sec:libero_plus}) bounds the structural-amplifier reading at the robot-initial-state factor. All experiments are simulation on LIBERO plus a single-architecture SimplerEnv sanity check. Sim-to-real transfer is open, under the class-level analog assumptions in \S\ref{sec:threat_mapping}. The stage-fused guard (\S\ref{sec:intervention}) is evaluated post hoc under an abstain-on-flag rule. Its adaptive evaluation covers an attacker against the action-anomaly term, not jointly with the text-consistency term. Future work can expand to closed-loop deployment and jointly adaptive evaluation.


\bibliographystyle{refstyle} 
\bibliography{references}

\newpage
\appendix
\begin{center}{\Large\bfseries Supplementary Material}\end{center}

\makeatletter
\let\app@section\section
\renewcommand{\section}{\FloatBarrier\app@section}
\makeatother

\section{Extended experimental details}
\label{app:details}

\paragraph{Compute.}
All experiments were conducted on A100 GPUs. Total compute: approximately 635 GPU-hours across 909 experiments (0 failures). FGSM gradient computation takes ${\sim}2.5\times$ longer than clean inference per episode; PGD-10 takes ${\sim}10\times$ longer. PGD-10 (Session 2c, 84 additional jobs) adds ${\sim}35$ GPU-hours. SimplerEnv cross-benchmark adversarial validation (Section~\ref{sec:simpler_env}; 120 jobs, OpenVLA-7B on 4 Google-Robot tasks $\times$ 3 seeds $\times$ 20 episodes across Gaussian, FGSM, PGD-10, and action-Gaussian attack families) adds ${\sim}40$ GPU-hours; the 660 prior LIBERO experiments account for the remaining ${\sim}520$ GPU-hours. Code for the gradient-checkpoint white-box PGD-10 attack through the $K{=}12$ latent recurrence, the EOT mechanism test, the consistency probe and adaptive evaluation harness, the $K$-sweep falsifier, and the LIBERO-90 within-paradigm sensitivity check will be \textbf{released upon publication}.

\paragraph{Evaluation protocol.}
Each condition was evaluated on all 4 LIBERO suites (10 tasks each) across 3 random seeds (42, 123, 456), yielding $N{=}12$ data points per condition. Each task was run for 5 episodes per seed, with a maximum of 300 steps per episode. Success was determined by LIBERO's built-in task completion check.

\paragraph{Model details.}
OpenVLA-OFT uses a Prismatic backbone with fine-tuned adapters and an MLPResNet action head (4 layers: $28672 \to 4096 \to 4096 \to 4096 \to 7$). DeepThinkVLA uses PaliGemma as its VLM backbone with a discrete action tokenizer (2,048 bins per dimension in $[-1, 1]$); CoT is generated via the language model head and action tokens are decoded by the same model in ``predict action from CoT'' mode. RD-VLA uses a Prismatic VLM backbone with 12 weight-tied recurrent transformer layers and a linear output projection ($7 \times 896$).

\paragraph{OpenVLA-family checkpoint disambiguation.}
Two distinct checkpoints of the OpenVLA architectural family appear in the paper; their absolute clean SRs are on different benchmarks and are not directly comparable. Table~\ref{tab:oft_variants} records the mapping; cross-benchmark verdicts (\S\ref{sec:simpler_env}) are read off $\Delta$SR within each row, not absolute SR across rows.
\begin{table}[h]
\centering
\small
\caption{OpenVLA-family checkpoints used in the paper.}
\label{tab:oft_variants}
\begin{tabularx}{\columnwidth}{@{}lllr>{\raggedright\arraybackslash}X@{}}
\toprule
Label & LIBERO FT & Used in & Clean SR & HuggingFace weights \\
\midrule
OpenVLA-OFT & yes & Tab.~\ref{tab:headline_3x3}, Fig.~1, \S\ref{sec:vulnerability} & $96.5\%$ (LIBERO, $N{=}12$) & \texttt{\seqsplit{moojink/openvla-7b-oft-finetuned-libero-spatial-object-goal-10}} \\
OpenVLA-7B & no & \S\ref{sec:simpler_env} & $55.0\%$ (SimplerEnv non-floor, $N{=}9$) & \texttt{\seqsplit{openvla/openvla-7b}} \\
\bottomrule
\end{tabularx}
\end{table}

\paragraph{FGSM implementation.}
For DeepThinkVLA, FGSM gradients are computed through the full vision-language pipeline. Because DeepThinkVLA emits discrete action tokens ($2{,}048$ bins per DOF), the attack loss is the cross-entropy between the model's action-token logits on the perturbed image and its clean greedy action-token prediction (used as a fixed target); maximizing this loss by gradient ascent on the image drives the predicted action tokens away from their clean values. The clean CoT is generated once and held fixed as prompt context while the perturbation is computed, then the full pipeline (including CoT re-generation) is re-run with the perturbed image.

\paragraph{$\varepsilon$-standardized cross-architecture comparison.}
\label{app:eps_std}
Table~\ref{tab:eps-std} pivots the existing white-box evaluations into a single budget-matched view: each row gives one architecture's clean SR plus FGSM and PGD-10 SR at $\varepsilon\in\{4/255,\,8/255\}$, all averaged over the same $N{=}12$ (4 suites $\times$ 3 seeds, 50 episodes/cell) protocol used in Table~\ref{tab:headline_3x3}. The $\varepsilon{=}8/255$ column reproduces the headline-table numbers; the $\varepsilon{=}4/255$ column is what justifies, with evidence rather than prose, our reporting at the canonical $8/255$ operating point. Three patterns hold at \emph{both} budgets, ruling out a $\varepsilon$-specific artifact: (i)~under FGSM, OFT is \emph{more} robust than DT ($83.8\%$ vs.\ $75.0\%$ at $4/255$; $83.5\%$ vs.\ $55.8\%$ at $8/255$), consistent with single-step gradient being too weak to expose the MLPResNet head; (ii)~under PGD-10, OFT collapses far below DT already at $\varepsilon{=}4/255$ ($35.2\%$ vs.\ $75.5\%$, a $\approx 40$~pp gap with non-overlapping $95\%$ CIs), and the gap widens at $8/255$ ($18.2\%$ vs.\ $49.8\%$); (iii)~under white-box PGD-10, RD-VLA collapses to $0.0\%$ at \emph{both} $\varepsilon{=}4/255$ and $\varepsilon{=}8/255$ across $4$ suites $\times$ $3$ seeds, decisively below OFT and DT and confirming that the architectural ranking under PGD-10 is DT $>$ OFT $>$ RD-VLA at every budget tested. The RD-VLA FGSM cells in this table remain DT$\to$RD-VLA transfer lower bounds (white-box FGSM on RD-VLA was not run because PGD-10 strictly dominates it); the white-box PGD-10 row is what fixes the cross-architecture ranking at the canonical operating points.
\begin{table}[t]
\centering
\small
\caption{$\varepsilon$-standardized vision robustness across the three architectures. Cells report mean LIBERO success rate (\%) $\pm$ half the 95\% bootstrap CI across N=12 (4 suites $\times$ 3 seeds, 50 episodes/cell). $\varepsilon$ values are matched across models within each column. RD-VLA PGD-10 cells are white-box via gradient checkpointing through the $K{=}12$ recurrence (\S\ref{sec:attacks}); RD-VLA FGSM cells ($^{\dagger}$) are DT$\to$RD-VLA transfer attacks providing a lower bound on white-box vulnerability (white-box FGSM was not run since PGD-10 strictly dominates it as the stronger attack on the same model).}
\label{tab:eps-std}
\resizebox{\ifdim\width>\textwidth \textwidth\else\width\fi}{!}{%
\begin{tabular}{lccccc}
\toprule
 & & \multicolumn{2}{c}{\textbf{FGSM}} & \multicolumn{2}{c}{\textbf{PGD-10}} \\
\cmidrule(lr){3-4}\cmidrule(lr){5-6}
Model & Clean & $\varepsilon{=}4/255$ & $\varepsilon{=}8/255$ & $\varepsilon{=}4/255$ & $\varepsilon{=}8/255$ \\
\midrule
OpenVLA-OFT & 96.5{\scriptsize$\,\pm\,$1.5} & 83.8{\scriptsize$\,\pm\,$12.4} & 83.5{\scriptsize$\,\pm\,$13.2} & 35.2{\scriptsize$\,\pm\,$12.0} & 18.2{\scriptsize$\,\pm\,$9.1} \\
DeepThinkVLA & 93.0{\scriptsize$\,\pm\,$4.2} & 75.0{\scriptsize$\,\pm\,$8.3} & 55.8{\scriptsize$\,\pm\,$12.5} & 75.5{\scriptsize$\,\pm\,$9.3} & 49.8{\scriptsize$\,\pm\,$11.2} \\
RD-VLA & 89.2{\scriptsize$\,\pm\,$5.7} & 87.7$^{\dagger}${\scriptsize$\,\pm\,$6.8} & 65.2$^{\dagger}${\scriptsize$\,\pm\,$20.5} & 0.0{\scriptsize$\,\pm\,$0.0} & 0.0{\scriptsize$\,\pm\,$0.0} \\
\bottomrule
\end{tabular}%
}
\end{table}

\paragraph{Continuous severity metric at the SR$=0$ floor.}
\label{app:e19_severity}
The two RD-VLA white-box PGD-10 cells in Table~\ref{tab:eps-std} both report $0.0\%$ SR, leaving open whether $\varepsilon{=}4/255$ is just past the collapse threshold or already deeply saturated. To disambiguate, we compute three continuous severity metrics on the same $24$ cells (no new GPU work) directly from the paired clean--PGD trajectories: mean per-step $\|\Delta\mathbf{a}\|_2$, max per-step $\|\Delta\mathbf{a}\|_2$, and cumulative end-effector drift (Table~\ref{tab:e19_severity}). All three metrics are statistically indistinguishable between $\varepsilon{=}4/255$ and $\varepsilon{=}8/255$ (relative change $\le 2\%$, within $\pm 1$~SE across $N{=}12$ cells), so $\varepsilon{=}4/255$ is not a marginal collapse but already at the same saturated regime as $\varepsilon{=}8/255$. This is consistent with the structural-amplification view of \S\ref{sec:vulnerability} (Prop.~\ref{prop:recurrence}): once perturbation magnitude exceeds the recurrence's basin of attraction, the per-iteration map drives outputs toward an $\varepsilon$-insensitive attractor, so the per-step action deviation $\rho \cdot \varepsilon$ does \emph{not} continue to scale with $\varepsilon$ in this range. Beyond closing the SR$=0$-vs-SR$=0$ ambiguity, this also bounds the diagnostic ceiling of any defense that monitors $\|\Delta\mathbf{a}\|_2$: at and above $\varepsilon{=}4/255$, action-magnitude alone cannot rank attack severity, so consistency-style probes (\S\ref{sec:adaptive_result}) that look at semantic alignment rather than action magnitude become the only remaining handle.
\begin{table}[t]
\centering
\caption{Continuous severity metrics for RD-VLA under white-box PGD-10 at the SR$=0$ floor. Mean per-step $\|\Delta\mathbf{a}\|_2$, max per-step $\|\Delta\mathbf{a}\|_2$, and cumulative end-effector drift over the episode are computed on paired clean--PGD trajectories and aggregated across $N{=}12$ cells ($4$ suites $\times$ $3$ seeds, $50$ episodes per cell). Both budgets produce SR$=0$, but the action-deviation magnitude is invariant to $\varepsilon$ in $[4,8]/255$ ($\le 2\%$ relative change, well within SE), consistent with the recurrence's output collapsing to an $\varepsilon$-insensitive attractor: $\varepsilon{=}4/255$ is therefore well past the collapse threshold rather than marginal.}
\label{tab:e19_severity}
\begin{tabular}{lrrrr}
\toprule
$\varepsilon$ & SR (\%) & $\overline{\|\Delta\mathbf{a}\|_2}$ & $\max\|\Delta\mathbf{a}\|_2$ & Cum.\ EEF drift (m) \\
\midrule
$4/255$ & 0.0 & 1.510 $\pm$ 0.016 & 2.566 $\pm$ 0.027 & 73.75 $\pm$ 13.96 \\
$8/255$ & 0.0 & 1.536 $\pm$ 0.009 & 2.591 $\pm$ 0.028 & 74.51 $\pm$ 13.60 \\
\bottomrule
\end{tabular}
\end{table}

\section{Consistency checker details}
\label{app:checker}

The consistency checker operates on a per-timestep basis and averages scores across the episode. Entity extraction uses the full LIBERO object vocabulary (29 objects) with longest-first substring matching. Directional keywords include: left, right, up, down, forward, backward, toward, away. Gripper intent is detected via keywords: open, close, grasp, release, grip, pick, place. Parseability is a binary check for sentence structure and minimum token count.

The threshold $\tau{=}0.7$ was selected from the threshold sweep: $\tau{=}0.5$ yields 84\% TPR / 0\% FPR; $\tau{=}0.6$ yields 93\% / 0\%; $\tau{=}0.7$ yields 100\% / 0\%; $\tau{=}0.8$ yields 100\% / 3\%; $\tau{=}0.9$ yields 100\% / 95\%.

\paragraph{Coherent-perturbation bound (formal statement).}
\label{app:coherent_bound}
\begin{observation}[Empirical separation bound on probe means, descriptive]\label{obs:coherent}
Let $\phi(\mathrm{CoT}, \mathbf{a})\in[0,1]$ be any consistency probe scoring plan-action agreement, and let $\bar{\phi}_\mathrm{clean}\!:=\!\mathbb{E}[\phi(f_r(\mathbf{z},\ell), f_a(f_r(\mathbf{z},\ell)))]$ measure its mean clean-data faithfulness. Vision-stage perturbations enter both branches through the shared perturbed latent $\tilde{\mathbf{z}}_t = f_v(\tilde{\mathbf{o}}_t)$, so $\widetilde{\mathrm{CoT}}$ and $\tilde{\mathbf{a}}$ are produced by the same composed map evaluated on $\tilde{\mathbf{z}}_t$. The mean clean-versus-attacked separation is bounded above by $1 - \bar{\phi}_\mathrm{clean}$. The bound is on means, not ROC; the empirical AUC numbers in Table~\ref{tab:detection} are the operative quantities.
\end{observation}

\paragraph{Bootstrap CI on $\bar\phi_\mathrm{clean}$ and Obs.~\ref{obs:coherent} sensitivity.}
\label{app:phi_clean_bootstrap}
The headline bound in \S\ref{sec:adaptive_result} uses the canonical $N{=}200$ clean-DT corpus from the vision-defense evaluation (\texttt{defense\_vision\_evaluation.json}: \texttt{libero\_object} seed $\in\{42,123,456\}$ + \texttt{libero\_spatial} seed$=42$): $\bar\phi_\mathrm{clean}{=}0.867 \pm 0.025$, $95\%$ bootstrap CI $[0.863, 0.870]$ ($10{,}000$ resamples), bound $1-\bar\phi_\mathrm{clean}{=}0.133$ ($95\%$ CI $[0.130, 0.137]$). To check the bound is not an artifact of those four conditions we additionally bootstrap on a broader $N{=}300$ pooled set covering all four LIBERO suites (Table~\ref{tab:phi_clean_per_suite}); the per-suite bounds range from $0.116$ to $0.145$, and the maximum measured clean-vs-attacked shift across our FGSM/Gaussian sweep ($0.022$, at FGSM $\varepsilon{=}16/255$) sits inside the $95\%$ CI of the bound for every suite individually.
\begin{table}[h]
\centering
\caption{Per-suite bootstrap of $\bar\phi_\mathrm{clean}$ and the implied empirical bound $1-\bar\phi_\mathrm{clean}$ (Obs.~\ref{obs:coherent}) on the broader $N{=}300$ pooled corpus. All bounds exceed the maximum measured $|\mathbb{E}[\phi]-\mathbb{E}[\tilde\phi]|{=}0.022$ across the FGSM/Gaussian vision-attack sweep.}
\label{tab:phi_clean_per_suite}
\begin{tabular}{lrcc}
\toprule
Suite & $N$ episodes & $\bar\phi_\mathrm{clean}$ ($95\%$ CI) & $1-\bar\phi_\mathrm{clean}$ ($95\%$ CI) \\
\midrule
\texttt{libero\_spatial} &  50 & $0.855$ $[0.826, 0.884]$ & $0.145$ $[0.116, 0.174]$ \\
\texttt{libero\_object}  & 100 & $0.875$ $[0.860, 0.889]$ & $0.125$ $[0.111, 0.140]$ \\
\texttt{libero\_goal}    & 100 & $0.864$ $[0.843, 0.883]$ & $0.136$ $[0.117, 0.157]$ \\
\texttt{libero\_10}      &  50 & $0.884$ $[0.861, 0.905]$ & $0.116$ $[0.095, 0.139]$ \\
\midrule
Pooled (broad)           & 300 & $0.869$ $[0.859, 0.879]$ & $0.131$ $[0.121, 0.141]$ \\
\midrule
Canonical defense-eval   & 200 & $0.867$ $[0.863, 0.870]$ & $0.133$ $[0.130, 0.137]$ \\
\bottomrule
\end{tabular}
\end{table}

\section{Adaptive attack implementation}
\label{app:adaptive}

Four adaptive attack variants of increasing sophistication:

\begin{enumerate}[leftmargin=*]
\item \textbf{Standard entity swap:} Replace CoT object mentions with random alternatives from the LIBERO vocabulary.
\item \textbf{Direction-adaptive:} Entity swap + match directional language in the CoT to the actual action vector direction (e.g., if action moves right, ensure CoT says ``right'').
\item \textbf{Stealth:} Entity swap + re-inject the original instruction objects into the corrupted CoT, fooling the entity-matching component.
\item \textbf{Full adaptive:} Combine direction fix and entity re-injection.
\end{enumerate}

The stealth variant reduces the entity component AUC from 1.000 to 0.492, demonstrating that entity re-injection alone is sufficient for complete evasion. Per-component AUC decomposition (Table~\ref{tab:component_auc_appendix}) confirms all detection power resides in the entity component.

\begin{table}[ht]
\centering
\caption{Per-component AUC decomposition under adaptive attacks.}
\label{tab:component_auc_appendix}
\begin{tabular}{@{}lccccc@{}}
\toprule
Attack & Entity & Direction & Gripper & Parse & Overall \\
\midrule
Standard entity swap & 1.000 & 0.500 & 0.500 & 0.500 & 0.996 \\
Direction-adaptive & 1.000 & 0.223 & 0.500 & 0.500 & 0.991 \\
Stealth & 0.492 & 0.500 & 0.500 & 0.482 & 0.493 \\
Full adaptive & 0.492 & 0.223 & 0.500 & 0.482 & 0.259 \\
\bottomrule
\end{tabular}
\end{table}

\section{Intervention I3: adaptive checker-aware PGD numerical table}
\label{app:I3_table}
Numerical values plotted in Figure~\ref{fig:I3_lambda_sweep} (\S\ref{sec:intervention}). Table~\ref{tab:intervention_I3_adaptive_vs_I2} reports per-cell raw success rate (undefended), defended success rate (abstain-on-flag), trigger rate, and leave-one-seed-out detection AUC; $n{=}150$ episodes per cell (3 seeds $\times$ 10 tasks $\times$ 5 episodes/task).
\begin{table}[t]
\centering
\small
\caption{Stage-fused guard ($\alpha{=}0.5$) under checker-aware adaptive PGD-10 ($\varepsilon{=}8/255$): numerical values for Figure~\ref{fig:I3_lambda_sweep}. $\lambda$ weights a differentiable action-anomaly penalty on the soft-argmaxed predicted action ($\lambda{=}0$ reproduces standard PGD-10). DT on \texttt{libero\_object} and \texttt{libero\_spatial}; $n{=}150$ episodes per cell (3 seeds $\times$ 10 tasks $\times$ 5 episodes/task); AUC is leave-one-seed-out detection AUC vs clean. ``Raw SR'' is undefended task success; ``Def. SR'' applies the abstain-on-flag rule; ``Trig.'' is the episode-level flag rate.}
\label{tab:intervention_I3_adaptive_vs_I2}
\resizebox{\ifdim\width>\textwidth \textwidth\else\width\fi}{!}{%
\begin{tabular}{llcccc}
\toprule
Suite & $\lambda$ & Raw SR (\%) & Def. SR (\%) & Trig. (\%) & AUC \\
\midrule
Object & 0.0 &  68.0{\scriptsize$\pm$7.3} &  59.3{\scriptsize$\pm$8.0} &  14.0{\scriptsize$\pm$5.7} & 0.895 \\
Object & 1.0 &  48.7{\scriptsize$\pm$8.0} &  46.0{\scriptsize$\pm$8.0} &  10.7{\scriptsize$\pm$5.0} & 0.717 \\
Object & 10.0 &  50.7{\scriptsize$\pm$8.0} &  46.0{\scriptsize$\pm$8.0} &   7.3{\scriptsize$\pm$4.3} & 0.669 \\
\midrule
Spatial & 0.0 &  64.7{\scriptsize$\pm$7.7} &  55.3{\scriptsize$\pm$8.0} &  28.0{\scriptsize$\pm$7.3} & 0.903 \\
Spatial & 1.0 &  64.0{\scriptsize$\pm$8.0} &  56.7{\scriptsize$\pm$8.0} &  22.0{\scriptsize$\pm$6.7} & 0.794 \\
Spatial & 10.0 &  60.7{\scriptsize$\pm$8.0} &  50.7{\scriptsize$\pm$8.0} &  30.7{\scriptsize$\pm$7.3} & 0.790 \\
\bottomrule
\end{tabular}%
}
\end{table}

\section{Stage-fused $\alpha$-sweep: per-suite breakdown}
\label{app:alpha_sweep}
The pooled $\alpha$-sweep (Table~\ref{tab:alpha_sweep}) is broken down per suite in Table~\ref{tab:alpha_sweep_per_suite}. The pooled $95\%$ CIs in Table~\ref{tab:alpha_sweep} are paired bootstraps ($n{=}10{,}000$ resamples, seed $0$) over the reconstructed per-episode Defended$-$Raw difference vector across all four suites ($N{=}600$). Per cell this difference is $-1$ exactly when a successful episode is flagged, so $\Delta \leq 0$ deterministically and the interval quantifies the magnitude of the shortfall. The clean and Gaussian rows show the abstain cost is present off the PGD-10 condition as well.

\begin{table}[t]
\centering
\small
\setlength{\tabcolsep}{4pt}
\caption{Per-suite $\Delta = $ Defended SR $-$ Raw SR (pp) under the $\alpha$-sweep, with paired-difference SE in subscript ($N{=}150$ per cell). Defended SR is bounded above by Raw SR under abstain-on-flag, so $\Delta \leq 0$ deterministically; values reported in pp. Trigger rates and per-cell Defended SR available in the auto-generated per-$\alpha$ tables \texttt{I2\_alpha\_*.tex}.}
\label{tab:alpha_sweep_per_suite}
\resizebox{\textwidth}{!}{%
\begin{tabular}{llcccccc}
\toprule
Suite & Attack & Raw SR (\%) & $\alpha{=}0$ & $\alpha{=}0.25$ & $\alpha{=}0.5$ & $\alpha{=}0.75$ & $\alpha{=}1$ \\
\midrule
Long & PGD-10 $\varepsilon{=}8/255$ & $18.0$ & $-1.3${\scriptsize$\pm0.9$} & $-4.7${\scriptsize$\pm1.7$} & $-2.7${\scriptsize$\pm1.3$} & $-1.3${\scriptsize$\pm0.9$} & $-0.7${\scriptsize$\pm0.7$} \\
Goal &  & $51.3$ & $-9.3${\scriptsize$\pm2.4$} & $-6.7${\scriptsize$\pm2.0$} & $-2.7${\scriptsize$\pm1.3$} & $-2.0${\scriptsize$\pm1.1$} & $-0.7${\scriptsize$\pm0.7$} \\
Object &  & $66.7$ & $-9.3${\scriptsize$\pm2.4$} & $-16.0${\scriptsize$\pm3.0$} & $-11.3${\scriptsize$\pm2.6$} & $-11.3${\scriptsize$\pm2.6$} & $+0.0${\scriptsize$\pm0.0$} \\
Spatial &  & $63.3$ & $-10.7${\scriptsize$\pm2.5$} & $-25.3${\scriptsize$\pm3.6$} & $-14.0${\scriptsize$\pm2.8$} & $-13.3${\scriptsize$\pm2.8$} & $-2.7${\scriptsize$\pm1.3$} \\
\midrule
Long & Gaussian $\sigma{=}0.2$ & $84.0$ & $-4.7${\scriptsize$\pm1.7$} & $-7.3${\scriptsize$\pm2.1$} & $-7.3${\scriptsize$\pm2.1$} & $-8.7${\scriptsize$\pm2.3$} & $-3.3${\scriptsize$\pm1.5$} \\
Goal &  & $94.0$ & $-3.3${\scriptsize$\pm1.5$} & $-5.3${\scriptsize$\pm1.8$} & $-4.0${\scriptsize$\pm1.6$} & $-4.0${\scriptsize$\pm1.6$} & $-0.7${\scriptsize$\pm0.7$} \\
Object &  & $96.7$ & $-4.0${\scriptsize$\pm1.6$} & $-8.0${\scriptsize$\pm2.2$} & $-8.7${\scriptsize$\pm2.3$} & $-10.0${\scriptsize$\pm2.4$} & $+0.0${\scriptsize$\pm0.0$} \\
Spatial &  & $96.0$ & $-5.3${\scriptsize$\pm1.8$} & $-10.7${\scriptsize$\pm2.5$} & $-6.7${\scriptsize$\pm2.0$} & $-9.3${\scriptsize$\pm2.4$} & $-6.0${\scriptsize$\pm1.9$} \\
\midrule
Long & Clean & $80.7$ & $-0.7${\scriptsize$\pm0.7$} & $-8.0${\scriptsize$\pm2.2$} & $-6.7${\scriptsize$\pm2.0$} & $-6.7${\scriptsize$\pm2.0$} & $-2.0${\scriptsize$\pm1.1$} \\
Goal &  & $96.0$ & $-1.3${\scriptsize$\pm0.9$} & $-6.7${\scriptsize$\pm2.0$} & $-4.0${\scriptsize$\pm1.6$} & $-2.7${\scriptsize$\pm1.3$} & $-2.0${\scriptsize$\pm1.1$} \\
Object &  & $98.7$ & $-4.0${\scriptsize$\pm1.6$} & $-10.0${\scriptsize$\pm2.4$} & $-10.7${\scriptsize$\pm2.5$} & $-10.7${\scriptsize$\pm2.5$} & $+0.0${\scriptsize$\pm0.0$} \\
Spatial &  & $96.7$ & $-5.3${\scriptsize$\pm1.8$} & $-9.3${\scriptsize$\pm2.4$} & $-10.0${\scriptsize$\pm2.4$} & $-11.3${\scriptsize$\pm2.6$} & $-8.7${\scriptsize$\pm2.3$} \\
\bottomrule
\end{tabular}%
}
\end{table}

\section{Lipschitz analysis details}
\label{app:lipschitz}

\paragraph{Spectral norm computation.}
Action head weights were extracted from model checkpoints and spectral norms computed via power iteration (100 iterations). For OFT's MLPResNet, the product-of-spectral-norms upper bound is:
\begin{equation}
L_\text{action} = \sigma(\text{fc1}) \cdot (1 + \sigma(\text{res0})) \cdot (1 + \sigma(\text{res1})) \cdot \sigma(\text{fc2}) = 22.67 \times 9.00 \times 9.15 \times 0.46 = 855.8
\end{equation}
This is a loose worst-case bound; the empirical end-to-end Lipschitz is only 0.56.

\begin{figure}[ht]
  \centering
  \includegraphics[width=0.55\textwidth]{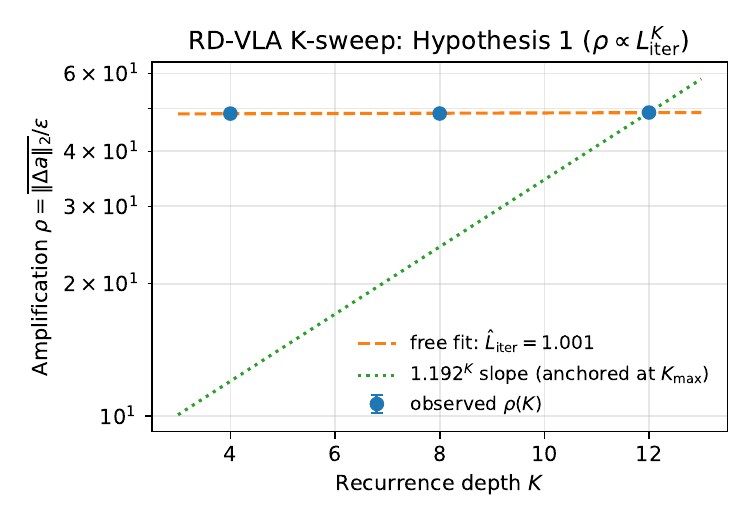}
  \caption{RD-VLA $K$-sweep at $\varepsilon{=}8/255$ ($N{=}12$ per $K$). Observed amplification $\rho(K)$ is statistically flat (slope gives $\hat{L}_\text{iter}{=}1.0007$, free-intercept fit), inconsistent with the back-solved $L_\text{iter}{=}1.192$ prediction (dotted line, anchored at $K{=}12$). The data positively identify a $K$-invariant ${\sim}49{\times}$ structural amplification rather than per-iteration compounding. \textbf{$K{=}4$ is out-of-distribution for this model (clean SR $72.8\%$ vs.\ $89\%$ at training $K{=}12$);} the multiplicative-per-iteration form is ruled out by the in-distribution $K{\in}\{8,12\}$ comparison alone.}
  \label{fig:e7_ksweep}
\end{figure}

\paragraph{Per-iteration Lipschitz derivation: back-solve and direct $K$-sweep test.}
From the empirical amplification data (Session~2f Phase~F paired sweep), $L_\text{reasoning}^\text{RD-VLA} / L_\text{reasoning}^\text{OFT} = 4.61 / 0.56 = 8.22$. Since OFT has no reasoning stage ($L_\text{reasoning}^\text{OFT} = 1.0$; its $L_\text{e2e} = 0.56$ reflects preprocessing attenuation in $L_\text{vision} \cdot L_\text{action}$), if the recurrence amplifies multiplicatively then $L_\text{reasoning}^\text{RD-VLA} = 8.22 = \hat{L}_\text{iter}^{12}$, giving $\hat{L}_\text{iter} = 8.22^{1/12} \approx 1.192$ ($\sim$19.2\% per iteration, compounding to $8.22\times$ at $K{=}12$). This back-solve is a one-parameter fit at a single $K$ and does not test the $L_\text{iter}^K$ form. We test it directly with an inference-time $K$-sweep on RD-VLA white-box PGD-10 at $\varepsilon{=}8/255$, varying $K{\in}\{4,8,12\}$ via the recurrence's $\texttt{num\_iter}$ flag (no retraining; 4 LIBERO suites $\times$ 3 seeds $\times$ 3 $K$ values = 36 cells, each paired with a clean baseline at the same $K$). Per-step amplification $\rho(K) = \overline{\|\Delta a\|_2}/\varepsilon$ is computed by pairing the clean and PGD trajectories within each (suite, seed, $K$) cell. Results (Table~\ref{tab:e7_ksweep}, Figure~\ref{fig:e7_ksweep}): $\rho(K{=}4){=}48.68\pm 0.62$, $\rho(K{=}8){=}48.70\pm 0.41$, $\rho(K{=}12){=}48.96\pm 0.30$. The pairwise ratio $\rho(K{=}12)/\rho(K{=}8){=}1.005$ (in-distribution $K$ values: clean SR $89\%$ vs.\ $88\%$) is inconsistent with the $\hat{L}_\text{iter}{=}1.192$ prediction of $1.192^4{\approx}2.02$. A free-intercept log-linear fit gives $\hat{L}_\text{iter}{=}1.0007$, mean abs.\ frac.\ error $0.1\%$. The $K{=}4$ cell carries an OOD-K caveat (clean SR $72.8\%$ vs.\ $89\%$ at $K{=}12$, since the model is trained at $K{=}12$ stochastic-recurrence-mean), but the $K{\in}\{8,12\}$ comparison alone is sufficient to rule out the multiplicative-per-iteration form on its own. \emph{Refined positive mechanism:} the architectural ${\sim}49{\times}$ amplification under PGD is real, $K$-invariant, and consistent with the cross-architecture amplification ordering OFT$<$DT$<$RD-VLA; the data localize the amplification to the encoder + first recurrence pass / fixed-point output, an interpretation consistent with the recurrence converging quickly to a fixed point such that further iterations neither add nor remove sensitivity. This is a sharper mechanistic claim than the original back-solve and identifies a more concrete intervention surface (smooth the encoder / fixed-point output) than reducing recurrence depth.

\section{Statistical tests}
\label{app:stats}

Table~\ref{tab:statistical_tests} reports the full set of cross-model and within-model significance tests used in Section~\ref{sec:vulnerability}. We use Shapiro--Wilk to select between parametric (paired $t$ / Welch's $t$) and non-parametric (Wilcoxon signed-rank / Mann--Whitney U) tests, and apply Holm--Bonferroni correction within this 13-test family ($\alpha{=}0.05$). Effect sizes are Cohen's $d$ (parametric) and Cliff's $\delta$ (rank-based). Sample size $N{=}12$ corresponds to 4 LIBERO suites $\times$ 3 seeds, with each data point being the mean SR across 50 episodes (10 tasks $\times$ 5 trials). OFT-row data are sourced from the Session~2f authors-eval-script v2 sweep; DT and RD-VLA rows are unchanged.

\begin{table}[t]
\centering
\caption{Statistical significance tests for key cross-model and within-model comparisons (Session~2f, OFT~v2 corrected). The three cross-model Gaussian $\sigma{=}0.2$ contrasts are paired by (suite,~seed) cell (matched $4\times3$ design); for every paired test the parametric/non-parametric choice is Shapiro--Wilk-selected on the paired differences. Holm--Bonferroni corrected within this 13-test family. Cohen's $d$ and Cliff's $\delta$ effect sizes reported. $*$: $p < 0.05$; $**$: $p < 0.01$; $***$: $p < 0.001$. Auto-generated by \texttt{scripts/build\_statistical\_tests\_table.py}; do not hand-edit.}
\label{tab:statistical_tests}
\resizebox{\textwidth}{!}{%
\begin{tabular}{lccccccc}
\toprule
Comparison & $N$ & Mean A & Mean B & Test & $p$ (corr.) & Cohen's $d$ & Cliff's $\delta$ \\
\midrule
  DT vs OFT, Gauss $\sigma{=}0.2$ & 12 & $92.7$ & $89.0$ & Wilcoxon & $0.3047$ & $0.87$ & $0.50$ \\
  OFT vs RD-VLA, Gauss $\sigma{=}0.2$ & 12 & $89.0$ & $14.8$ & Wilcoxon & $0.0049$** & $6.58$ & $1.00$ \\
  DT vs RD-VLA, Gauss $\sigma{=}0.2$ & 12 & $92.7$ & $14.8$ & paired $t$ & $<\!0.001$*** & $6.50$ & $1.00$ \\
  DT clean vs Gauss $\sigma{=}0.2$ & 13 & $93.1$ & $93.2$ & paired $t$ & $0.8858$ & $-0.02$ & $0.07$ \\
  RD clean vs Gauss $\sigma{=}0.2$ & 12 & $89.2$ & $14.8$ & Wilcoxon & $0.0044$** & $5.52$ & $1.00$ \\
  OFT clean vs Gauss $\sigma{=}0.2$ & 12 & $96.5$ & $89.0$ & Wilcoxon & $0.0039$** & $3.65$ & $1.00$ \\
  DT action $\sigma{=}0.1$ vs $0.5$ & 12 & $86.7$ & $8.0$ & paired $t$ & $<\!0.001$*** & $7.09$ & $1.00$ \\
  RD action $\sigma{=}0.1$ vs $0.5$ & 12 & $76.2$ & $8.7$ & paired $t$ & $<\!0.001$*** & $4.92$ & $1.00$ \\
  OFT action $\sigma{=}0.1$ vs $0.5$ & 12 & $80.5$ & $0.0$ & Wilcoxon & $0.0034$** & $12.78$ & $1.00$ \\
  OFT clean vs FGSM $\varepsilon{=}2/255$ & 12 & $96.5$ & $96.2$ & paired $t$ & $1.0000$ & $0.12$ & $0.08$ \\
  OFT clean vs FGSM $\varepsilon{=}8/255$ & 12 & $96.5$ & $83.5$ & Wilcoxon & $0.0078$** & $0.68$ & $0.52$ \\
  OFT clean vs PGD-10 $\varepsilon{=}8/255$ & 12 & $96.5$ & $18.2$ & Wilcoxon & $0.0029$** & $6.43$ & $1.00$ \\
  OFT FGSM vs PGD-10 $\varepsilon{=}8/255$ & 12 & $83.5$ & $18.2$ & Wilcoxon & $0.0049$** & $2.90$ & $0.83$ \\
\bottomrule
\end{tabular}%
}
\end{table}

The DT vs.\ OFT comparison under Gaussian $\sigma{=}0.2$ is not significant after correction ($p{=}0.30$, $d{=}0.87$): the pp-level gap (3.7) is small and within seed-to-seed variance for DT. The full three-tier ordering is nevertheless preserved with high confidence by joint bootstrap (P(DT $>$ OFT $>$ RD-VLA)$=0.997$ over 10{,}000 resamples), driven by the consistent direction of the small DT--OFT gap and the very large gap to RD-VLA. The headline OFT-specific findings---FGSM resistance ($d{=}0.12$ at $\varepsilon{=}2/255$, $d{=}0.68$ at $\varepsilon{=}8/255$) versus PGD-10 vulnerability ($d{=}6.43$ at $\varepsilon{=}8/255$, $p{<}0.01$ corrected) and the FGSM-vs-PGD-10 gap ($d{=}2.90$, $p{<}0.01$ corrected)---survive correction with very large effect sizes.

\section{Per-suite breakdown}
\label{app:per_suite}

\begin{table}[t]
\centering
\caption{Cross-stage vulnerability profile. SR (\%) mean $\pm$ SE across 4 LIBERO suites and 3 seeds ($N{=}12$). FGSM: DT white-box; OFT white-box (Session 2b); RD-VLA via DT$\to$RD-VLA transfer (lower bound, ${}^t$). \textbf{Reasoning-stage attacks (entity-swap CoT) apply to DT only}: RD-VLA has no text CoT and OFT has no reasoning module, so the corresponding cells are structurally N/A and not a coverage gap.}
\label{tab:cross_stage}
\small
\resizebox{\ifdim\width>\textwidth \textwidth\else\width\fi}{!}{%
\begin{tabular}{llccc}
\toprule
Stage & Attack & OpenVLA-OFT & DeepThinkVLA & RD-VLA \\
\midrule
  --- & Clean & $96.5 \pm 0.8$ & $93.0 \pm 2.1$ & $89.2 \pm 2.9$ \\
  Vision & Gaussian $\sigma{=}0.01$ & $97.0 \pm 0.3$ & $91.5 \pm 2.2$ & $87.5 \pm 4.4$ \\
  Vision & Gaussian $\sigma{=}0.05$ & $90.2 \pm 5.3$ & $93.3 \pm 1.8$ & $73.8 \pm 8.4$ \\
  Vision & Gaussian $\sigma{=}0.1$ & $96.0 \pm 0.4$ & $93.3 \pm 2.4$ & $54.2 \pm 8.9$ \\
  Vision & Gaussian $\sigma{=}0.2$ & $89.0 \pm 0.3$ & $92.7 \pm 1.6$ & $14.8 \pm 4.4$ \\
  Vision & FGSM $\varepsilon{=}2/255$ & $96.2 \pm 0.9$ & $86.2 \pm 3.1$ & $88.0 \pm 3.9$$^t$ \\
  Vision & FGSM $\varepsilon{=}4/255$ & $83.8 \pm 7.1$ & $75.0 \pm 4.3$ & $87.7 \pm 3.5$$^t$ \\
  Vision & FGSM $\varepsilon{=}8/255$ & $83.5 \pm 7.8$ & $55.8 \pm 6.4$ & $65.2 \pm 10.8$$^t$ \\
  Vision & FGSM $\varepsilon{=}16/255$ & $84.5 \pm 2.9$ & $41.5 \pm 5.4$ & $36.2 \pm 6.7$$^t$ \\
  Vision & FGSM $\varepsilon{=}32/255$ & $78.8 \pm 3.6$ & $25.8 \pm 4.3$ & $7.3 \pm 2.3$$^t$ \\
\midrule
  Action & Action $\sigma{=}0.01$ & $96.0 \pm 0.6$ & $92.2 \pm 2.4$ & $86.7 \pm 4.0$ \\
  Action & Action $\sigma{=}0.05$ & $91.5 \pm 1.5$ & $90.8 \pm 2.7$ & $86.7 \pm 4.3$ \\
  Action & Action $\sigma{=}0.1$ & $80.5 \pm 2.5$ & $86.7 \pm 3.8$ & $76.2 \pm 4.9$ \\
  Action & Action $\sigma{=}0.5$ & $0.0 \pm 0.0$ & $8.0 \pm 2.0$ & $8.7 \pm 2.3$ \\
  Action & Action $\sigma{=}1.0$ & $1.5 \pm 0.8$ & $0.8 \pm 0.4$ & $1.0 \pm 0.6$ \\
\midrule
  Reasoning & Entity swap (CoT) & N/A & $87.0$ & N/A \\
  Reasoning & Shuffled (CoT) & N/A & $95.5$ & N/A \\
\bottomrule
\end{tabular}%
}
\end{table}

Table~\ref{tab:cross_stage} gives the full cross-stage SR profile (mean${\pm}$SE per model and attack), and the cross-model and within-model significance tests behind \S\ref{sec:vulnerability} are reported in App.~\ref{app:stats}. Under Gaussian noise at $\sigma{=}0.2$, the three-tier pattern is consistent across all suites, with amplified contrast on libero\_object:

\begin{table}[ht]
\centering
\caption{Per-suite SR (\%) under Gaussian noise $\sigma{=}0.2$ (mean of 3 seeds).}
\begin{tabular}{@{}lccccc@{}}
\toprule
Model & Object & Spatial & Goal & Long & Overall \\
\midrule
DeepThinkVLA & 96.7 & 92.0 & 92.7 & 89.3 & 92.7 \\
RD-VLA & 0.0 & 19.3 & 22.0 & 18.0 & 14.8 \\
OpenVLA-OFT & 90.0 & 88.0 & 88.0 & 90.0 & 89.0 \\
\bottomrule
\end{tabular}
\end{table}

RD-VLA achieves 0\% SR across all 3 seeds on libero\_object, the most extreme demonstration of latent reasoning amplification.

\section{Confound analysis (extended)}
\label{app:confounds}

The three models under study differ in multiple dimensions beyond reasoning type, and we must carefully assess which observed effects are attributable to reasoning versus other architectural factors.

\paragraph{Confounding variables.}
Four factors vary across our models: (1)~\emph{backbone}: Prismatic (OpenVLA-OFT, RD-VLA) vs.\ PaliGemma (DeepThinkVLA); (2)~\emph{parameter count}: ${\sim}$7B vs.\ ${\sim}$3B vs.\ 0.5B; (3)~\emph{action head}: MLPResNet vs.\ discrete tokens (2,048~bins) vs.\ linear projection; (4)~\emph{training data}: different fine-tuning procedures and data augmentation.

\paragraph{Partial backbone control.}
RD-VLA and OFT share the Prismatic VLM backbone family, isolating reasoning type (latent iterative vs.\ none) from backbone effects. Under Gaussian $\sigma{=}0.2$, OFT degrades $-7.5$pp while RD-VLA collapses $-74.3$pp---a $9.9\times$ ratio that cannot be explained by backbone differences. In normalized terms (controlling for different clean baselines), relative SR drops are 7.8\% (OFT) vs.\ 83.9\% (RD-VLA), confirming the effect survives baseline normalization.

\paragraph{Parameter count as potential confounder.}
The most vulnerable model (RD-VLA, 0.5B) is also the smallest; amplification could partially reflect scale-dependent representation fragility rather than reasoning alone. However, the 7B OFT model (14$\times$ larger) shows \emph{intermediate} SR-vulnerability but \emph{lowest} amplification ratio (0.56), with its preprocessing attenuating Gaussian noise more than DT (1.84) or RD-VLA (4.61). DT (3B, intermediate scale) shows the least SR vulnerability despite intermediate amplification---because its CoT reasoning provides semantic error correction. The SR-vulnerability ordering (DT~$<$~OFT~$<$~RD-VLA) does not align with either the parameter ordering (RD-VLA~$<$~DT~$<$~OFT) or the amplification ordering (OFT~$<$~DT~$<$~RD-VLA), suggesting that reasoning type (specifically, the combination of spatial attenuation and semantic correction) is the dominant factor, not model scale.

\paragraph{Action head effects.}
RD-VLA's linear output projection is a contraction ($\sigma_1{=}0.091$), ruling out the action head as the amplifier. DT's discrete tokenization provides an implicit quantization barrier. The amplification originates in the recurrent reasoning layers, not downstream.

\paragraph{Direct ablation: CoT-disabled DT.}
We run DT with its reasoning chain suppressed (zero tokens between \texttt{<think>} and \texttt{</think>}), holding backbone, scale, action head, and training constant (full results in Fig.~\ref{fig:cot_ablation} and Tab.~\ref{tab:cot_ablation}, App.~\ref{app:cot_ablation}). Under $\sigma{=}0.2$ noise, CoT-disabled degrades by $1.5$~pp (92.7\%$\to$91.2\%) while CoT-enabled degrades by only $0.3$~pp; the gap is in the expected direction but does not reach significance at $N{=}12$: a paired $t$ gives $p{=}0.35$ (the value cited in the main text) and the Shapiro--Wilk-selected Wilcoxon signed-rank test on the same non-normal paired differences gives $p{=}0.59$ ($d{=}0.28$); both agree the CoT-on advantage is undetected at this sample size. Clean SR is unchanged ($p{=}0.81$, $d{=}0.07$). \emph{Power analysis:} at $N{=}12$ with $\alpha{=}0.05$ and $80\%$ power, the minimum detectable effect size is $d{\approx}0.85$ (Cohen's $d$, two-sample $t$). The observed $d{=}0.28$ falls below this detection threshold; an $N{\geq}120$ design would be required to detect the observed effect at standard power. Following~\citet{lakens2018equivalence}, two complementary null-bounding analyses are appropriate here. (i) A TOST equivalence test at $\pm 5$pp boundaries on $\Delta_{\text{CoT}}$ at $\sigma{=}0.2$ confirms equivalence at $\alpha{=}0.05$ ($p_\text{TOST}{=}0.021$ for Gaussian, $p_\text{TOST}{=}0.003$ for clean). (ii) The JZS Bayes factor under a default Cauchy prior (scale $0.707$) gives BF$_{01}{=}2.32$ (Gaussian) and BF$_{01}{=}3.39$ (clean), anecdotal-to-substantial evidence for the null. Per-episode pairing of the existing trajectories (suite$\times$seed$\times$task$\times$episode-index, $N{=}700$ paired episodes for Gaussian, $N{=}650$ for clean) shrinks the minimum detectable effect from Cohen $d{=}0.85$ to Cohen $h{=}0.106$ (roughly a $5$pp difference in raw SR), with TOST $\pm 5$pp confirmed at $p{<}0.001$ for both conditions, $\Delta\mathrm{SR}{=}{+}1.43$pp [95\% CI $-0.59,+3.45$] for Gaussian, and a McNemar discordant count of $b{=}31/c{=}21$ ($p{=}0.21$, two-sided exact). We therefore frame the result as: at this power level we cannot detect a CoT contribution to noise robustness; we do not claim CoT contributes nothing, and the equivalence/Bayes-factor reanalyses will bound how large any undetected effect can be.

\section{EOT mechanism check (full table)}
\label{app:eot_mechanism}
Table~\ref{tab:e9_eot} reports the EOT~$T{=}2$ result discussed in \S\ref{sec:eot_mechanism}: EOT narrows the FGSM-vs-PGD-10 gap to ${\leq}4$pp on 3/4 LIBERO suites, and across suites strengthens PGD-10 on average ($\Delta_\text{PGD}{=}{-}1.7$pp), ruling out gradient masking as the explanation for the within-OFT FGSM-vs-PGD asymmetry.
\begin{table}[t]
\centering
\small
\setlength{\tabcolsep}{4pt}
\caption{EOT (T=2) mechanism test on OFT (LIBERO-fine-tuned OpenVLA-OFT). Each cell is mean SR \% $\pm$ SE over 3 seeds $\times$ 50 episodes. Augmentation set: $T{=}2$ samples per gradient step with $\pm 2$~px translation jitter and $\pm 0.02$ brightness perturbation. \textbf{Headline ($\varepsilon{=}8/255$):} FGSM-EOT vs.\ FGSM and PGD-10-EOT vs.\ PGD-10 deltas. Mean $\Delta_\text{PGD}{=}{-}1.7$pp across all 4 suites; $|\Delta|{\leq}4$pp on 3/4 suites for both attacks; gradient masking is refuted at the canonical operating point. \textbf{Descriptive ($\varepsilon{=}16/255$):} EOT and PGD-10-EOT supplementary measurements; PGD-10-EOT saturates to $0.0\%$ on all 4 LIBERO suites, consistent with the recurrence-saturation regime of App.~\ref{app:e19_severity}. The libero\_10 FGSM cell at $\varepsilon{=}8/255$ has $\sigma{=}49$pp on $n{=}3$ (one seed flip); the PGD-10-EOT delta on the same suite ($-4.0$pp) is clean and consistent with the other three suites, and is what carries the per-suite verdict.}
\label{tab:e9_eot}
\resizebox{\textwidth}{!}{%
\begin{tabular}{l|cccc|cc}
\toprule
\multicolumn{1}{l|}{} & \multicolumn{4}{c|}{Headline ($\varepsilon{=}8/255$)} & \multicolumn{2}{c}{Descriptive ($\varepsilon{=}16/255$)} \\
Suite & FGSM & FGSM-EOT & PGD-10 & PGD-10-EOT & FGSM-EOT & PGD-10-EOT \\
\midrule
libero\_object  & $99.3 \pm 1.2$ & $98.0 \pm 2.0$ & $40.0 \pm 24.6$ & $40.7 \pm 1.2$ & $96.7 \pm 1.2$ & $0.0 \pm 0.0$ \\
libero\_spatial & $86.7 \pm 2.3$ & $82.7 \pm 1.2$ & $10.0 \pm 2.0$ & $\phantom{0}8.0 \pm 3.5$ & $83.3 \pm 1.2$ & $0.0 \pm 0.0$ \\
libero\_goal    & $91.3 \pm 4.2$ & $94.0 \pm 4.0$ & $11.3 \pm 5.0$ & $10.0 \pm 5.3$ & $85.3 \pm 3.1$ & $\phantom{0}0.0 \pm 0.0$ \\
libero\_10      & $56.7 \pm 49.1^{\dagger}$ & $86.0 \pm 5.3$ & $11.3 \pm 2.3$ & $\phantom{0}7.3 \pm 1.2$ & $78.7 \pm 4.2$ & $\phantom{0}0.0 \pm 0.0$ \\
\midrule
$\Delta$ (mean) & --- & $\phantom{-}{+}6.7^{\dagger}$ & --- & $-1.7$ & --- & --- \\
\bottomrule
\end{tabular}%
}
\\[2pt]
\footnotesize $^{\dagger}$libero\_10 FGSM at $\varepsilon{=}8/255$: $\sigma{=}49$pp on $n{=}3$ seeds (one seed-flip outlier dominates the mean); the PGD-10-EOT $\Delta{=}{-}4.0$pp on the same suite is clean and consistent with the other three suites, and is the per-suite signal we read mechanism off. Excluding libero\_10 FGSM, mean $\Delta_\text{FGSM}{=}{-}0.9$pp across the 3 clean suites.
\end{table}

\section{Amplification analysis: full results}
\label{app:amplification}

\begin{figure}[t]
  \centering
  \begin{subfigure}[t]{0.24\textwidth}
    \includegraphics[width=\textwidth]{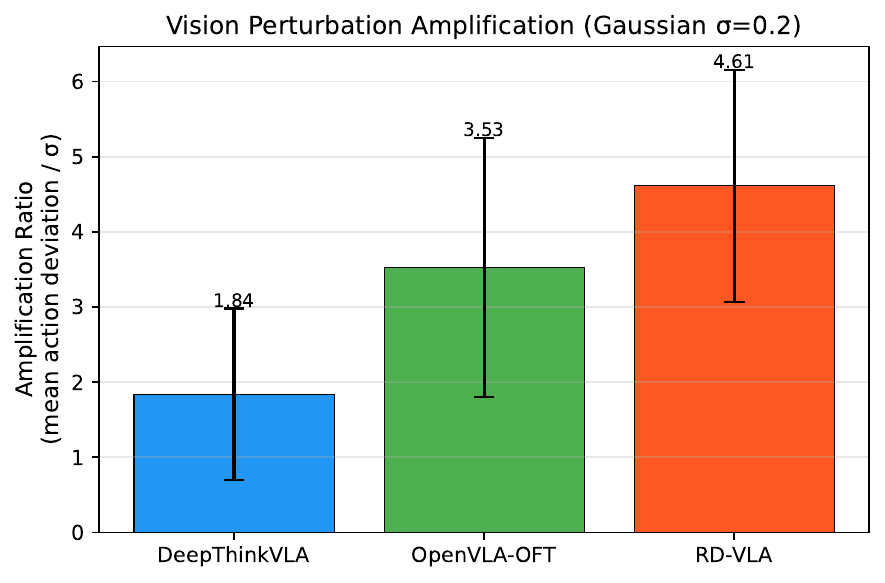}
    \caption{Amplification ratios at $\sigma{=}0.2$}
  \end{subfigure}\hfill
  \begin{subfigure}[t]{0.24\textwidth}
    \includegraphics[width=\textwidth]{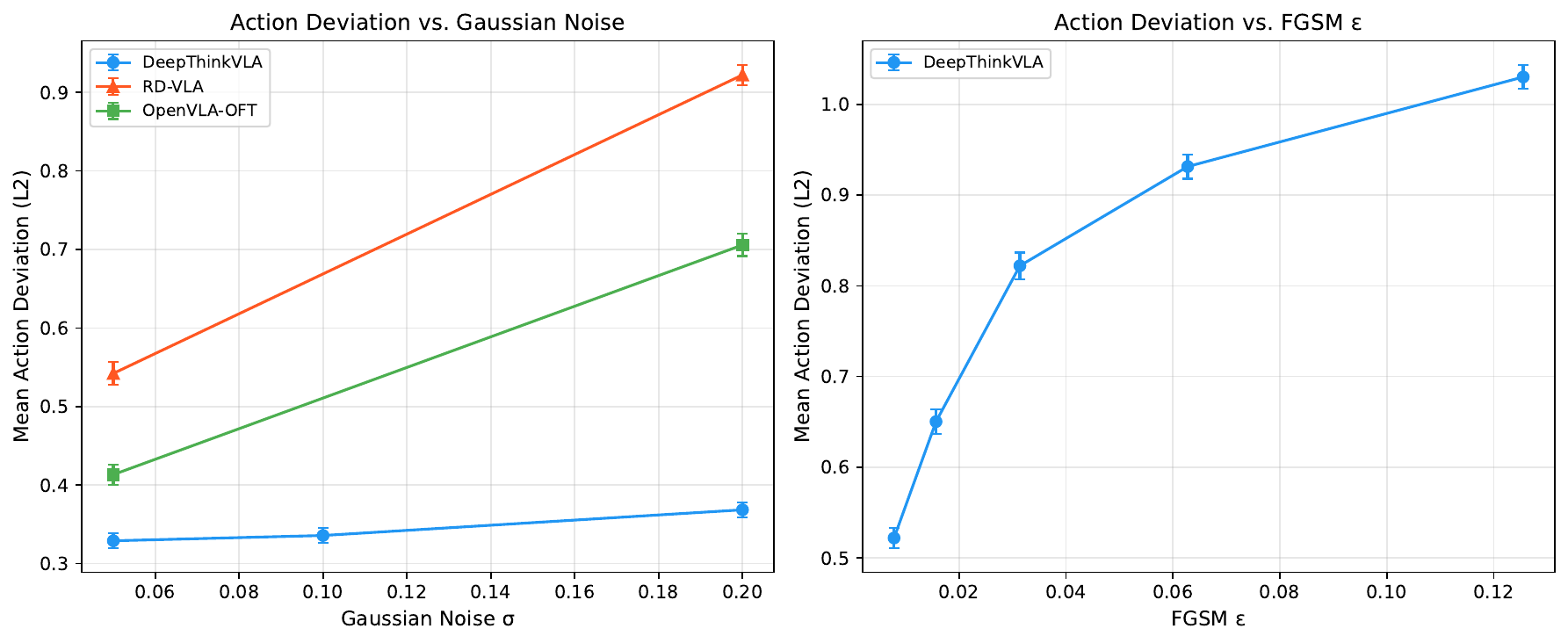}
    \caption{SR dose-response}
  \end{subfigure}\hfill
  \begin{subfigure}[t]{0.24\textwidth}
    \includegraphics[width=\textwidth]{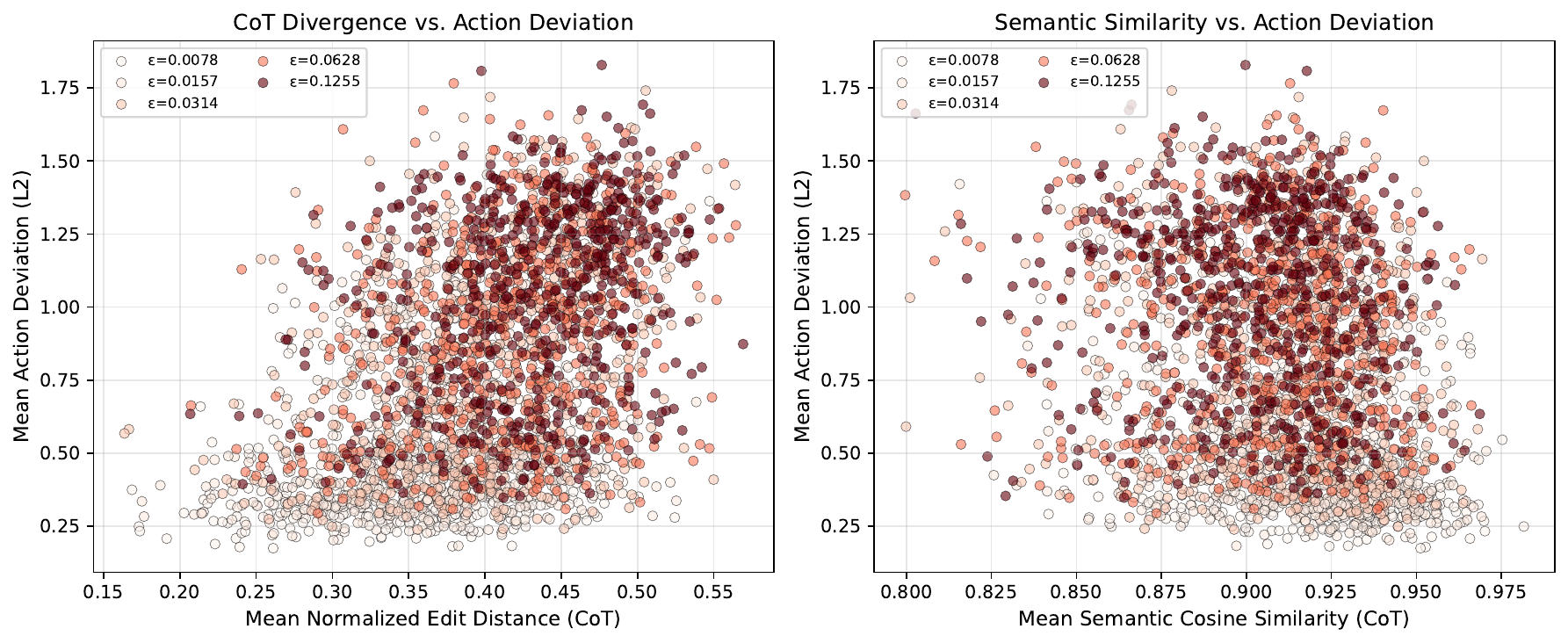}
    \caption{CoT cascade ($r{=}{+}0.44$)}
  \end{subfigure}\hfill
  \begin{subfigure}[t]{0.24\textwidth}
    \includegraphics[width=\textwidth]{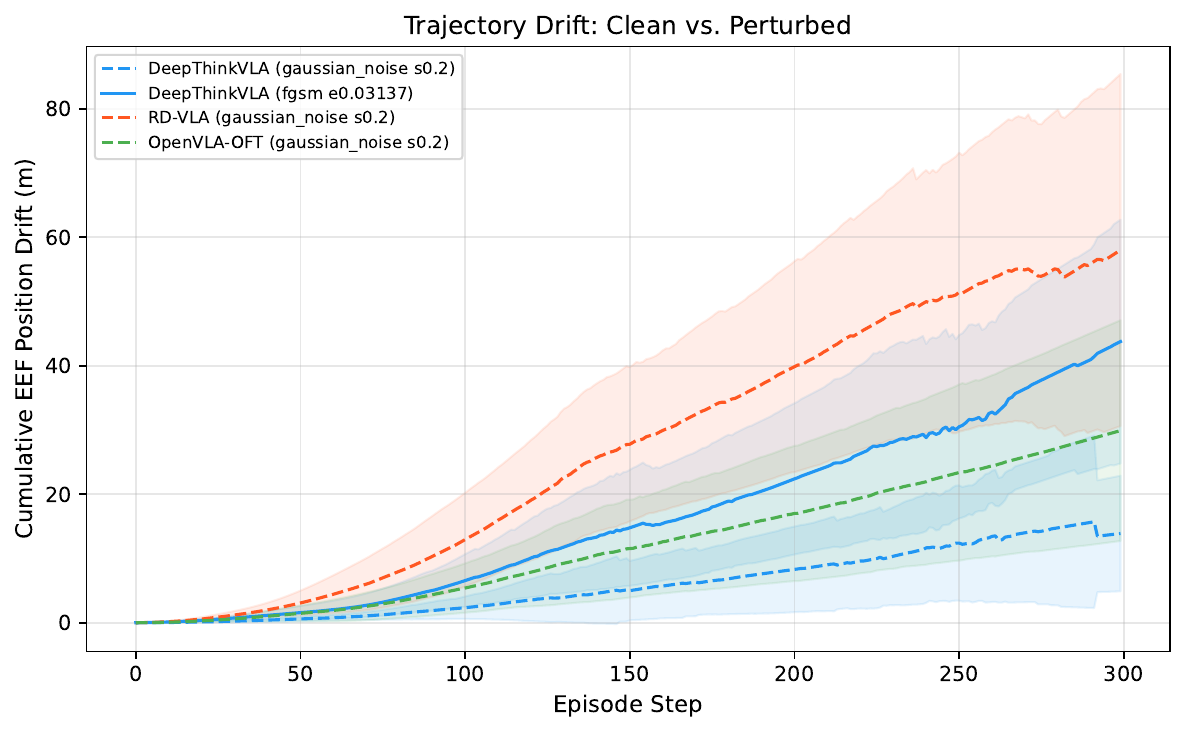}
    \caption{Trajectory drift}
  \end{subfigure}
  \caption{Amplification cascade ($7{,}752$ paired clean--perturbed episode records). OFT attenuates ($\rho_\text{OFT}{\approx}0.56$), DT amplifies moderately ($\rho_\text{DT}{\approx}1.84$), RD-VLA amplifies strongly ($\rho_\text{RD-VLA}{\approx}4.61$, $2.5{\times}$ DT). The recurrent reasoning stage is the structural multiplier ($8.22{\times}$ over the no-reasoning baseline; $K$-invariant per Prop.~\ref{prop:recurrence}).}
  \label{fig:amplification}
\end{figure}

Table~\ref{tab:amplification} reports per-model, per-attack amplification ratios alongside CoT divergence, and Figure~\ref{fig:amplification} visualizes the full cascade.

\begin{table}[t]
\centering
\caption{Amplification analysis across models and attack types. Amplification ratio = mean action deviation / perturbation magnitude. CoT divergence measured for DeepThinkVLA under FGSM (text reasoning enables measurement). SR $\leftrightarrow$ action deviation correlation across all episodes.}
\label{tab:amplification}
\resizebox{\textwidth}{!}{%
\begin{tabular}{lcccccc}
\toprule
Model & Attack & Amp.\ Ratio & Action Dev & CoT Edit Dist & Sem.\ Cosine & SR$\downarrow$ \\
\midrule
  OpenVLA-OFT & Gauss $\sigma{=}0.2$ & $0.56 \pm 0.19$ & $0.112$ & --- & --- & $89.0\%$\textsuperscript{*} \\
  DeepThinkVLA & Gauss $\sigma{=}0.2$ & $1.84 \pm 1.14$ & $0.368$ & --- & --- & $92.7\%$ \\
  RD-VLA & Gauss $\sigma{=}0.2$ & $4.61 \pm 1.55$ & $0.922$ & --- & --- & $14.8\%$ \\
\midrule
  DeepThinkVLA & FGSM $\varepsilon{=}2/255$ & $66.58$ & $0.522$ & $0.366$ & $0.918$ & $86.2\%$ \\
  DeepThinkVLA & FGSM $\varepsilon{=}4/255$ & $41.45$ & $0.650$ & $0.378$ & $0.916$ & $75.0\%$ \\
  DeepThinkVLA & FGSM $\varepsilon{=}8/255$ & $26.21$ & $0.822$ & $0.401$ & $0.910$ & $55.8\%$ \\
  DeepThinkVLA & FGSM $\varepsilon{=}16/255$ & $14.84$ & $0.931$ & $0.415$ & $0.906$ & $41.5\%$ \\
  DeepThinkVLA & FGSM $\varepsilon{=}32/255$ & $8.21$ & $1.030$ & $0.429$ & $0.903$ & $25.8\%$ \\
\midrule
  \multicolumn{7}{l}{\textit{SR $\leftrightarrow$ action deviation: Pearson $r = -0.668$, $p < 0.0001$ ($N = 7752$)}} \\
  \multicolumn{7}{l}{\textsuperscript{*}OFT SR from Session~2f v2 sweep (authors' original eval script; 50 ep/condition)} \\
\bottomrule
\end{tabular}%
}
\end{table}

\section{Per-stage propagation bounds}
\label{app:propagation_bounds}

\begin{figure}[ht]
  \centering
  \includegraphics[width=0.85\textwidth]{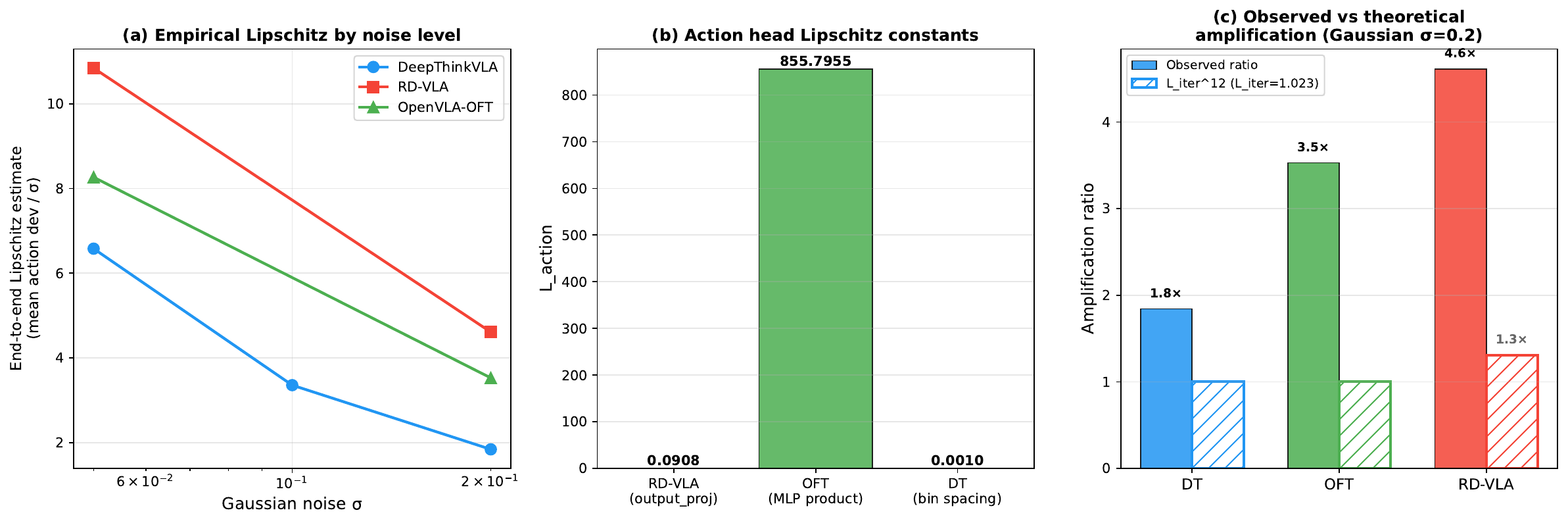}
  \caption{Perturbation amplification analysis. \textbf{Left:} per-stage spectral norms---RD-VLA's output projection is a contraction ($\sigma_1{=}0.091$); amplification occurs upstream in recurrent layers. \textbf{Center:} end-to-end amplification at $\sigma{=}0.2$. \textbf{Right:} the $K$-sweep test at $K{\in}\{4,8,12\}$ (Figure~\ref{fig:e7_ksweep}, Table~\ref{tab:e7_ksweep}) finds $\rho(K)$ statistically flat ($\hat{L}_\text{iter}{\approx}1.0$): the architectural ${\sim}49{\times}$ amplification is positively identified as $K$-invariant and structural (encoder + fixed-point recurrence output). The back-solved $\hat{L}_\text{iter} \approx 1.192$ (from the observed $K{=}12$ ratio $L_\text{reasoning}^\text{RD-VLA}/L_\text{reasoning}^\text{OFT} \approx 8.22$) implies a multiplicative-per-iteration form that the $K$-sweep rules out.}
  \label{fig:propagation}
\end{figure}

\begin{figure}[ht]
  \centering
  \includegraphics[width=0.55\linewidth]{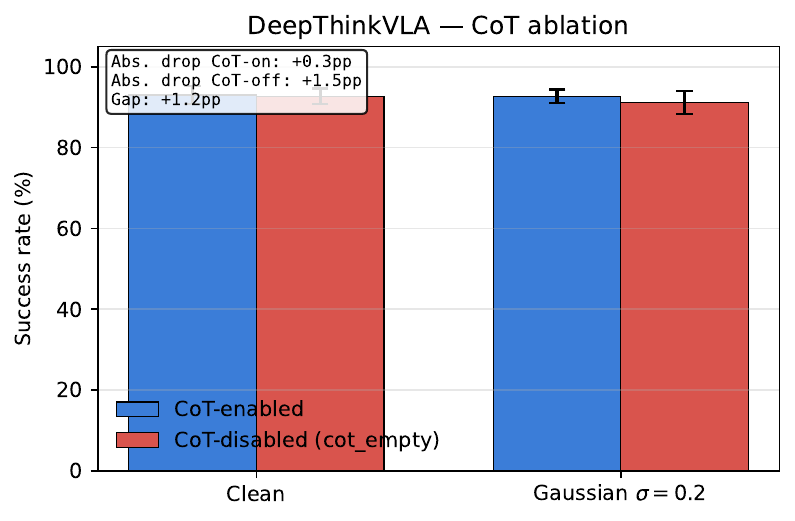}
  \caption{CoT-disabled ablation on DeepThinkVLA ($N{=}12$ suite/seed pairs). Bars show mean SR ($\pm$SE) under clean and Gaussian $\sigma{=}0.2$ conditions. CoT-enabled drops $0.3$~pp under noise; CoT-disabled drops $1.5$~pp (gap $+1.2$~pp; Shapiro-selected Wilcoxon $p{=}0.59$, paired $t$ $p{=}0.35$, both non-significant; $d{=}0.28$). Clean SR is indistinguishable ($p{=}0.81$). Results indicate backbone-driven robustness, not CoT-driven denoising.}
  \label{fig:cot_ablation}
\end{figure}

\paragraph{Formal definitions.}
The end-to-end \emph{amplification ratio} is
\begin{equation}
  \rho \;=\; \frac{\overline{\|\mathbf{a}_\text{pert} - \mathbf{a}_\text{clean}\|_2}}{\varepsilon},
  \label{eq:amplification}
\end{equation}
the per-step action deviation per unit input perturbation under a stage-wise perturbation (\S\ref{sec:threat_model}). The Lipschitz upper bound on end-to-end deviation through the three-stage pipeline (Eq.~\ref{eq:pipeline}) is
\begin{equation}
  \|\boldsymbol{\delta}_a\|_2 \;\leq\; L_\text{action} \cdot L_\text{reasoning} \cdot L_\text{vision} \cdot \|\boldsymbol{\delta}_v\|_2,
  \label{eq:perstage}
\end{equation}
and for weight-tied recurrence $f_r = g^{\circ K}$ the multiplicative hypothesis specialises $L_\text{reasoning} = L_\text{iter}^K$. Figure~\ref{fig:propagation} summarizes the per-stage spectral norms and end-to-end bound, Table~\ref{tab:propagation_bounds} the per-stage Lipschitz estimates and bound tightness, and the full spectral-norm computation and back-solve are in App.~\ref{app:lipschitz}.

\begin{table}[t]
\centering
\caption{Per-stage Lipschitz estimates and bound tightness under Gaussian noise ($\sigma{=}0.2$). $L_\text{action}$ is computed via spectral norm for continuous action heads (RD-VLA output projection, OFT MLP) or bin spacing for discrete tokenization (DT). $L_\text{reasoning}$ is estimated empirically relative to OFT (no reasoning baseline). $\hat{L}_\text{iter}$ is the per-iteration Lipschitz constant derived from the compound analysis $\hat{L}_\text{iter}^{12} = L_\text{reasoning}$.}
\label{tab:propagation_bounds}
\small
\resizebox{\ifdim\width>\textwidth \textwidth\else\width\fi}{!}{%
\begin{tabular}{@{}lccccccc@{}}
\toprule
Model & Reasoning & $L_\text{action}$ & $L_\text{reasoning}$ & $\hat{L}_\text{iter}$ & $L_\text{e2e}$ & Obs.\ $\bar{\delta}_a$ & Amp.\ ratio \\
\midrule
OpenVLA-OFT & None & 855.80\textsuperscript{c} & $<1.0$\textsuperscript{e} & --- & 0.56 & 0.112 & 0.56 \\
DeepThinkVLA & Text CoT & 0.0010\textsuperscript{a} & $\approx 1.0$\textsuperscript{b} & --- & 1.84 & 0.368 & 1.84 \\
RD-VLA & Recurrent (12$\times$) & 0.0908\textsuperscript{d} & 8.22 & 1.192 & 4.61 & 0.922 & 4.61 \\
\bottomrule
\end{tabular}%
}

\vspace{2pt}
{\footnotesize
\textsuperscript{a}Discrete action tokenization with 2048 bins; spacing = max per-DOF change from 1-bin shift. \\
\textsuperscript{b}Text CoT acts as stochastic denoiser under Gaussian noise (observed $\Delta$SR $\approx 0$pp). \\
\textsuperscript{c}Product of per-layer spectral norms through MLPResNet: $\sigma(\text{fc1})=22.67 \times (1+\sigma(\text{res0}))=9.00 \times (1+\sigma(\text{res1}))=9.15 \times \sigma(\text{fc2})=0.46$. \\
\textsuperscript{d}Spectral norm of final linear projection (7$\times$896 matrix). \\
\textsuperscript{e}OFT's dual-backbone preprocessing (SigLIP + DINOv2) attenuates Gaussian noise: $L_\text{e2e} = 0.56 < 1$. $L_\text{reasoning}^\text{RD-VLA}$ is estimated as $L_\text{e2e}^\text{RD-VLA} / L_\text{e2e}^\text{OFT} = 4.61/0.56 = 8.22$ (Phase~F paired sweep, 1,800 episode records).
}
\end{table}

\section{CoT-disabled ablation: full results}
\label{app:cot_ablation}

\begin{table}[t]
\centering
\caption{CoT-disabled ablation on DeepThinkVLA (LIBERO, $N{=}12$ paired suite/seed pairs). Success rates (\%, mean $\pm$ SE) under clean and Gaussian $\sigma{=}0.2$ vision perturbation. ``CoT-disabled'' replaces the reasoning chain with empty CoT. The \emph{gap} column is the difference in absolute drop between the two variants; a positive value means removing CoT increases noise sensitivity. Paired Wilcoxon signed-rank test on the Gaussian condition (Shapiro-selected for the non-normal paired differences): $p{=}0.5938$; a paired $t$ on the same differences gives $p{=}0.35$, both non-significant. Paired Cohen's $d{=}0.28$.}
\label{tab:cot_ablation}
\begin{tabular}{lccc}
\toprule
Variant & Clean SR & Gauss $\sigma{=}0.2$ SR & Absolute drop \\
\midrule
  CoT-enabled & $93.0 \pm 2.2$ & $92.7 \pm 1.7$ & $+0.3$ pp \\
  CoT-disabled (empty CoT) & $92.7 \pm 1.9$ & $91.2 \pm 2.8$ & $+1.5$ pp \\
\midrule
  \textbf{Gap (disabled $-$ enabled)} & --- & --- & $\mathbf{+1.2}$ \textbf{pp} \\
\bottomrule
\end{tabular}
\end{table}

\section{SimplerEnv cross-benchmark: per-task details and degenerate case}
\label{app:simpler_env}

\paragraph{Setup.}
We use OpenVLA-7B~\citep{kim2024openvla} (no-reasoning, OpenVLA class) evaluated via its publicly-released checkpoint without fine-tuning. We evaluate four Google-Robot tasks (\texttt{google\_robot\_pick\_coke\_can}, \texttt{move\_near}, \texttt{open\_drawer}, \texttt{close\_drawer}) at 3 seeds $\times$ 20 episodes per condition, using \texttt{fractal20220817\_data} action normalization. WidowX excluded ($\sim 1\%$ clean SR). ECoT-Bridge~\citep{zawalski2024ecot} was attempted but lacks the required normalization key and achieves $0\%$ on WidowX with $\sim 300$s/episode, making comparative evaluation infeasible.

\paragraph{Per-task FGSM heterogeneity.}
FGSM $\varepsilon{=}8/255$ $\Delta$SR by task: \texttt{close\_drawer} $-5$pp, \texttt{move\_near} $+6.7$pp, \texttt{open\_drawer} $-35$pp. PGD-10 is uniformly severe: $-20$, $-20$, $-45$pp respectively. The PGD$>$FGSM ordering holds on every individual task.

\paragraph{Floor-dominated case: \texttt{pick\_coke\_can}.}
This task achieves only $10\%$ clean SR (vs.\ published $30$--$60\%$~\citep{li2024simpler}), indicating a systematic failure mode in the public OpenVLA-7B checkpoint. For floor-dominated baselines, adversarial attacks paradoxically \emph{increase} SR ($+28$pp under FGSM, $+33$pp under PGD-10) by disrupting the failure mode---a known property of adversarial examples near the performance floor. We exclude it from cross-benchmark comparisons.

\section{LIBERO-Plus naturalistic-perturbation cross-benchmark: per-(model, factor) details}
\label{app:libero_plus}

This appendix gives the full per-(model, factor) breakdown behind the main-text cross-class corroboration (\S\ref{sec:libero_plus}); the pre-registered 16-task task selection is the canonical sample, and the 64-task layout supplementary is parallel evidence on task-selection sensitivity rather than a replacement of the canonical cell.

\paragraph{Setup.}
We evaluate the three reasoning paradigms (DT, RD-VLA, OpenVLA-OFT) on LIBERO-Plus~\citep{fei2025liberoplus}'s seven naturalistic-perturbation factors. The pre-registered task set (\texttt{experiments/libero\_plus/tasks.json}; numpy seed${=}0$ for the original four factors, per-factor seeds $\{1, 2, 3\}$ for the Amendment-1 additions) draws 16 tasks per factor, stratified across the four LIBERO suites. Each (model, factor) cell pools $3$ seeds $\times$ 16 tasks $\times$ 5 episodes ($n{=}240$). All three models use their LIBERO-fine-tuned checkpoints from the main paper (\S\ref{sec:threat_model}); LIBERO-Plus applies its naturalistic perturbations upstream of the model adapters, so all three see identical perturbations within a factor.

\begin{table}[ht]
\centering
\small
\caption{LIBERO-Plus~\citep{fei2025liberoplus} per-(model, factor) success rate (\%) and paired RD-VLA-vs-\{DT, OFT\} $\Delta$SR gap. Each cell pools $3$ seeds $\times$ 16 tasks $\times$ 5 episodes ($n{=}240$) except the supplementary objects-layout row at $n{=}960$ ($3$ seeds $\times$ 64 tasks $\times$ 5 ep). Pre-registered task lists at \texttt{experiments/libero\_plus/tasks.json}; the supplementary layout sample at \texttt{layout\_supplementary\_tasks.json} (independent numpy seed${=}42$). Paired-gap columns are $\Delta\mathrm{SR}_{\mathrm{RD-VLA}} - \Delta\mathrm{SR}_{\mathrm{other}}$ with $\Delta\mathrm{SR}$ measured against the same-model clean-baseline row. The RD-vs-OFT gap clears $30$\,pp on six of seven factors (the 64-task supplementary entry resolves the 16-task layout sampling artifact, \S\ref{sec:libero_plus}); robot-initial-state is the exception at $+9.6$\,pp.}
\label{tab:libero_plus_main}
\resizebox{\textwidth}{!}{%
\begin{tabular}{l c c c c c}
\toprule
factor & DT & OFT & RD-VLA & gap RD-DT & gap RD-OFT \\
\midrule
clean baseline (factor=none) & 96.2 & 95.0 & 92.5 & --- & --- \\
\midrule
camera viewpoint & 76.2 & 57.5 & 13.8 & $+58.8$ & $+41.2$ \\
robot initial state & 47.5 & 37.5 & 25.4 & $+18.3$ & $+9.6$ \\
sensor noise & 88.8 & 78.8 & 15.0 & $+70.0$ & $+61.2$ \\
language instructions & 72.5 & 85.0 & 25.4 & $+43.3$ & $+57.1$ \\
objects layout (16-task primary) & 60.0 & 46.2 & 20.0 & $+36.2$ & $+23.8$ \\
\quad{}objects layout (64-task suppl.) & 72.8 & 66.2 & 25.3 & $+43.8$ & $+38.4$ \\
\midrule
light conditions & 90.0 & 97.5 & 41.7 & $+44.6$ & $+53.3$ \\
background textures & 76.2 & 77.5 & 12.1 & $+60.4$ & $+62.9$ \\
\bottomrule
\end{tabular}%
}
\end{table}

\paragraph{Per-factor pattern.}
The RD-VLA-vs-\{DT, OFT\} paired $\Delta$SR gap exceeds $30$\,pp on both contrasts for camera viewpoint, sensor noise, language instructions, light conditions, background textures, and (via the 64-task supplementary) objects layout---six of seven factors. The exception is robot initial state, where the RD-vs-OFT gap is only $+9.6$\,pp: under starting-state variance OFT drops from $95.0\%$ clean to $37.5\%$ while RD-VLA drops from $92.5\%$ to $25.4\%$. The recurrence-driven amplifier story (\S\ref{sec:ksweep_result}, Prop.~\ref{prop:recurrence}) does not extend to initial-state perturbations: starting-state variance mis-grounds all three policies' visual--object correspondence approximately equally, leaving little headroom for RD-VLA's K${=}12$ recurrence to differentially amplify. The OFT anchor against LIBERO-Plus's published leaderboard passes the pre-registered $\pm 10$\,pp tolerance on 4 of 5 high-severity factors (camera $-2.2$, robot init $+0.3$, sensor noise $+2.0$, language $+3.5$\,pp); objects layout is the single anchor failure ($-30.9$\,pp on the 16-task primary sample), diagnosed below.

\paragraph{Objects-layout selection-bias diagnostic.}
The pre-registered 16-task objects-layout sub-sample is bimodally distributed: 8 of 16 sampled tasks score OFT${=}0\%$ and 6 of 16 score $100\%$, with mean per-task difficulty $3.25$ vs the LIBERO-Plus pool mean $2.89$ (the seed${=}1$ stratified draw oversampled difficulty levels $3$--$5$). An independently-seeded 64-task supplementary (\texttt{experiments/libero\_plus/layout\_supplementary\_tasks.json}; numpy seed${=}42$; $n{=}960$ episodes per model) at a pool-matched difficulty distribution (mean $2.92$) produced OFT~$=66.2\%$ (16-task primary: $46.2\%$; LIBERO-Plus published 250-task aggregate: $77.1\%$), closing the anchor gap from $-30.9$ to $-10.9$\,pp and the paired RD-vs-OFT gap from $+23.8$ to $+38.4$\,pp. The remaining $11$\,pp deficit vs the published 250-task aggregate is consistent with the supplementary sample still being smaller than the full pool. The supplementary is reported as parallel evidence on layout-factor task-selection sensitivity; the pre-registered 16-task results in Tab.~\ref{tab:libero_plus_main} are unmodified, and the cross-stage two-tier ordering on layout follows the supplementary numbers (gap RD-vs-OFT $+38.4$\,pp $\geq 30$\,pp); for the layout factor, the pre-registered 16-task cell is canonical and the 64-task supplementary is reported as parallel evidence on layout-factor task-selection sensitivity, not as a re-classification of the pre-registered result. Per-cell raw data: \texttt{code/data/libero\_plus\_full\_aggregate.csv}.

\section{Limitations (extended)}
\label{app:limitations}

We deliberately focus on a precise, controlled comparison across three representative reasoning-architecture models (text CoT, latent iterative, none; one per family) on a single well-established manipulation benchmark; this scope makes within-benchmark statistical comparisons well-powered ($N{=}12$ paired suite/seed cells; $N{\geq}650$ paired episodes for the within-DT CoT-disabled ablation) but bounds the immediate generalization claims, which are model-level observations rather than paradigm-level claims. (1)~We study two reasoning-architecture families with a single model each; visual reasoning VLAs (e.g., image-plan-predicting models) are an interesting third axis we do not yet cover. (2)~SimplerEnv (\S\ref{sec:simpler_env}) replicates the FGSM/PGD asymmetry on a different robot. (3)~The consistency probe uses keyword-level entity matching; semantic-level analysis would shift but not eliminate the structural coherently-wrong gap (Obs.~\ref{obs:coherent}). (4)~RD-VLA PGD-10 is white-box via gradient checkpointing through the $K{=}12$ recurrence and Prismatic LLM (peak $3.5$\,GB at $K{=}12$ on A100-40GB; \S\ref{sec:attacks}); RD-VLA FGSM is by DT$\to$RD-VLA transfer (a lower bound) since PGD-10 strictly subsumes FGSM as the stronger attack on the same model; black-box Square Attack on RD-VLA directly serves as a gradient-free cross-check. (5)~Baseline performance varies across architectures ($93\%/89\%/96.5\%$); we emphasize relative degradation alongside absolute comparisons. (6)~We attempted to add a second text-CoT VLA (ECoT~\citep{zawalski2024ecot}) via LoRA fine-tuning of the public ECoT-bridge checkpoint on $2{,}000$ LIBERO trajectories; despite fixing three inference bugs and an upstream LoRA-scaling clamp (\texttt{lora\_alpha=min(rank,16)} giving scaling$=0.25$ vs the standard $1$--$2$), three orthogonal recipe interventions (12K-step training, $8\times$ LoRA scaling, $50\times$ action-token loss reweighting) all converged to the same $\approx 0.45$ action-accuracy plateau and $0\%$ smoke SR. We diagnose this as a representational-capacity limit of LoRA on the Bridge-pretrained base under Bridge$\to$LIBERO action distribution shift. Full fine-tune~\citep{chen2025ecotlite} ($76.6\%$ on LIBERO-90) exceeds our $\sim$100 GPU-hours/run budget plus the attack sweep. The architectural CoT-vs-no-reasoning contrast is independently corroborated by the latent-vs-no-reasoning contrast within our three-model panel. The within-DT CoT-disabled ablation's control sets the reasoning text to zero tokens, off-manifold relative to an in-distribution skip-CoT comparator, so its $\pm 5$pp equivalence bound is a conservative reading of the CoT effect rather than a tight one (full ablation in App.~\ref{app:confounds}).

Three narrower limitations are deferred here from the main text. \emph{Reasoning-stage corruption is text-CoT-only by architectural availability:} latent reasoning (RD-VLA) has no addressable reasoning surface and single-pass (OFT) has no reasoning stage, so the reasoning-stage column of the cross-stage matrix has a single populated cell. \emph{The probe evaluation covers a single probe class:} the adaptive consistency-probe evaluation tests entity-matching plan-action consistency on text CoT; other probe classes (sentence-embedding similarity, learned consistency classifiers, semantic-role parsers) might shift the AUC ceiling but are not evaluated here. \emph{The EOT mechanism test uses $T{=}2$:} the test ruling out gradient masking for OFT's FGSM/PGD asymmetry uses $T{=}2$ stochastic augmentations per gradient step, below the literature norm $T\in\{8,16\}$; the curvature-on-the-action-head conclusion is cross-checked with the gradient-free Square Attack at 100 queries but is not airtight against obfuscated-gradient skepticism, and a higher-$T$ EOT run plus direct Hessian-trace measurement (E21) are deferred.



\end{document}